\documentclass{article}

\usepackage{microtype}
\usepackage{graphicx}
\usepackage{subfigure}
\usepackage{booktabs} 

\usepackage{hyperref}

\usepackage[accepted]{icml2023}

\usepackage{amsmath}
\usepackage{amssymb}
\usepackage{mathtools}
\usepackage{amsthm}

\usepackage[capitalize,noabbrev]{cleveref}

\theoremstyle{plain}
\newtheorem{theorem}{Theorem}[section]

\newtheorem{lemma}[theorem]{Lemma}
\newtheorem{corollary}[theorem]{Corollary}
\theoremstyle{definition}

\theoremstyle{remark}

\usepackage{graphicx}
\usepackage{float} 
\usepackage{import}
\usepackage{wrapfig,lipsum,booktabs}
\usepackage{caption}
\usepackage{multirow}
\usepackage{xcolor}

\usepackage{xspace}

\def\ie{\emph{i.e.}\xspace}

\newcommand{\goto}{\rightarrow}

\usepackage[textsize=tiny]{todonotes}

\icmltitlerunning{Vector Quantized Wasserstein Auto-Encoder}

\begin{document}

\twocolumn[
\icmltitle{Vector Quantized Wasserstein Auto-Encoder}



\icmlsetsymbol{equal}{*}

\begin{icmlauthorlist}
\icmlauthor{Tung-Long Vuong}{yyy,xxx}
\icmlauthor{Trung Le}{yyy}
\icmlauthor{He Zhao}{comp}
\icmlauthor{Chuanxia Zheng}{sch}
\icmlauthor{Mehrtash Harandi}{yyy}
\icmlauthor{Jianfei Cai}{yyy}
\icmlauthor{Dinh Phung}{yyy,xxx}

\end{icmlauthorlist}

\icmlaffiliation{yyy}{Monash University, Australia}
\icmlaffiliation{xxx}{Vinai, Vietnam}
\icmlaffiliation{comp}{CSIRO's Data61, Australia}
\icmlaffiliation{sch}{University of Oxford, United Kingdom}

\icmlcorrespondingauthor{Tung-Long Vuong}{Tung-Long.Vuong@monash.edu}

\icmlkeywords{Machine Learning, ICML}

\vskip 0.3in
]



\printAffiliationsAndNotice{}  

\begin{abstract}
Learning deep discrete latent presentations offers a promise of better symbolic and summarized abstractions that are more useful to subsequent downstream tasks. Inspired by the seminal Vector Quantized Variational Auto-Encoder (VQ-VAE), most of work in learning deep discrete representations has mainly focused on improving the original VQ-VAE form and none of them has studied learning deep discrete representations from the generative viewpoint. In this work, we study learning deep discrete representations from the generative viewpoint. Specifically, we endow discrete distributions over sequences of codewords and learn a deterministic decoder that transports the distribution over the sequences of codewords  to the data distribution via minimizing a WS distance between them. We develop further theories to connect it with the clustering viewpoint of WS distance, allowing us to have a better and more controllable clustering solution. Finally, we empirically evaluate our method on several well-known benchmarks, where it achieves better qualitative and quantitative performances than the other VQ-VAE variants in terms of the codebook utilization and image reconstruction/generation.
\end{abstract}

\section{Introduction}
Learning compact yet expressive representations from large-scale and high-dimensional unlabeled data is an important and long-standing task in machine learning \citep{kingma2013auto, chen2020simple,chen2021exploring}. Among many different kinds of methods, Variational Auto-Encoder (VAE) \citep{kingma2013auto} and its variants \citep{tolstikhin2017wasserstein,alemi2016deep,higgins2016beta,voloshynovskiy2019information} have shown great success in unsupervised representation learning.
Although these continuous representation learning methods have been successfully applied to various problems, ranging from images \citep{pathak2016context, goodfellow2014generative, kingma2016improved}, video, and audio \citep{reed2017parallel, oord2016wavenet, kalchbrenner2017video}, in some contexts, input data is more naturally modeled and encoded as discrete symbols rather than continuous ones. For example, discrete representations are a natural fit for complex reasoning, planning, and predictive learning \citep{van2017neural}. This motivates the need for learning discrete representations while preserving the insightful characteristics of the input data. The Vector Quantization Variational Auto-Encoder (VQ-VAE) \citep{van2017neural} is a pioneering generative model that successfully combines the VAE framework with discrete latent representations. In particular, vector quantized models learn a compact discrete representation using a deterministic encoder-decoder architecture in the first stage and subsequently apply this highly compressed representation to various downstream tasks. Examples include image generation \citep{esser2021taming}, cross-modal translation \citep{kim2022verse}, and image recognition \citep{yu2021vector}. 

VQ-VAE aims at learning encoder-decoder and a trainable codebook. The \emph{codebook} is formed by set of codewords $C=\{c_{k}\}_{k=1}^{K}$
on the latent space $\mathcal{Z} \in \mathbb{R}^{n_z}$ ($C\in \mathbb{R}^{K \times n_z}$). We denote a $M$-dimensional discrete latent space related to the codebook  as the $M$-ary Cartesian power of $C$: $ C^{M} \in \mathbb{R}^{M \times n_z}$  with $M$ is the number of components in the latent space. 
We also denote a latent variable in  $C^{M}$ and its $m$-th component as $\Bar{z}_n \in C^{M}$ and $\Bar{z}_{n}^{m} \in C$ respectively.
The \emph{encoder} $f_{e}: \mathbb{R}^{n_x}\rightarrow \mathbb{R}^{M \times n_z}$ first
map the data examples $x_n\in \mathbb{R}^{n_x}$ to the latent $z_n \in \mathbb{R}^{M \times n_z}$ ($z_{n}^{m}=f_{e}^{m}(x_n)$ is the $m$-th component of $z_n$), followed by a quantization $Q_C$ projecting $z_n$ onto $C^{M}:\Bar{z_n}=Q_C(z_n)$. 
The quantization process is modelled as
a deterministic categorical posterior distribution such that: $\Bar{z}_{n}^m=\text{argmin} _{k}\rho_{z}\left(f_{e}^{m}\left(x_n\right),c_{k}\right)$ where $\rho_{z}$ is a metric on the latent space. 
The \emph{decoder} $f_{d}:  \mathbb{R}^{M  \times n_z} \rightarrow \mathbb{R}^{n_x}$ reconstructs accurately the data examples from the discrete latent representations.

The objective function of VQ-VAE is as follows:
\begin{equation*}
\mathbb{E}_{x\sim\mathbb{P}_{x}}
\begin{bmatrix}
[d_{x}\left(f_{d}\left(Q_C({f}_{e}\left(x\right))\right),x\right)
 \\
+d_{z}\left(\text{\textbf{sg}}\left(f_{e}\left(x\right)\right),\Bar{z}\right)+\beta d_{z}\left(f_{e}\left(x\right),\text{\textbf{sg}}\left(\Bar{z}\right)\right)
\end{bmatrix},
\end{equation*}
where $\mathbb{P}_{x}=\frac{1}{N}\sum_{n=1}^{N}\delta_{x_{n}}$ is
the empirical data distribution, \textbf{sg} specifies stop gradient,
$d_{x}$ is a distance on data space, and $\beta$ is set between $0.1$ and $2.0$ \citep{van2017neural}. 


While VQ-VAE has been widely applied to representation learning in many areas \citep{henter2018deep,baevski2020wav2vec,razavi2019generating,kumar2019melgan, dieleman2018challenge, yan2021videogpt,hu2022unified}, it is known to suffer from \emph{codebook collapse}, which has a low codebook usage, \ie most of embedded latent vectors are quantized to just few discrete codewords, while the \emph{other codewords are rarely used, or dead}. This issue arises due to the poor initialization of the codebook, which reduces the information capacity of the bottleneck \citep{roy2018theory,takida22a,yu2021vector}. 

To mitigate this issue, several additional training heuristics were proposed, such as the \textit{exponential moving average} (EMA) update \citep{van2017neural,razavi2019generating}, soft expectation maximization (EM) update \citep{roy2018theory}, codebook reset \citep{dhariwal2020jukebox, williams2020hierarchical}. Notably, the \textit{soft expectation maximization} (EM) update \citep{roy2018theory} connects the EMA update with an EM algorithm and softens the EM algorithm with a stochastic posterior. \textit{Codebook reset} randomly reinitializes unused or low-used codewords to one of the encoder outputs \citep{dhariwal2020jukebox} or those near codewords of high usage \citep{williams2020hierarchical}. \citet{takida22a} extends the standard VAE by incorporating stochastic quantization and a trainable posterior categorical distribution. Their findings demonstrate that annealing the stochasticity of the quantization process leads to a significant improvement in codebook utilization.

Recently, Wasserstein (WS) distance has been applied successfully to \textit{generative models} and \textit{\underline{continuous} representation learning} \citep{arjovsky17a, NIPS2017_892c3b1c, tolstikhin2017wasserstein} owing to its nice properties and theory. It is natural to ask: \textit{"Can we take advantages of intuitive properties of the WS distance and its mature theory for learning compact yet expressive \underline{discrete} representations?"} 

Towards addressing this question, in this paper, we develop solid theories by connecting the theory bodies and viewpoints of the WS distance, generative models, and deep discrete representation learning. In particular, we establish theories for the real and practical setting of learning discrete representation in which a data example $\mathbf{X}$ is mapped to a sequence of $M$ latent codes $\mathbf{Z}= [\mathbf{Z}^1, \dots, \mathbf{Z}^M]$ corresponding to a sequence of $M$ codewords $\mathbf{C}= [\mathbf{C}^1, \dots, \mathbf{C}^M]$ via an encoder $f_e$. Our theory development pathway is as follows. We first endow $M$ discrete distributions over $\mathbf{C}^1,\dots,\mathbf{C}^M$, sharing a common support set as the set of codewords $C= [c_k]_{k=1}^K \in \mathbb{R}^{K \times n_z}$. We then use a joint distribution $\gamma$, admitting these discrete distributions over $\mathbf{C}^1,\dots,\mathbf{C}^M$ as its marginal distributions to sample a sequence of $M$ codewords $\mathbf{C}= [\mathbf{C}^1, \dots, \mathbf{C}^M]$. From the generative 
 viewpoint, we propose learning a decoder $f_d$ to minimize the \textit{codebook-data distortion} as the WS distance: $\mathcal{W}_{d_z}(f_d\#\gamma, \mathbb{P}_x)$ (cf. (\ref{eq:push_forward})). 

 Subsequently, we develop rigorous theories to equivalently turn the formulation in the generative viewpoint to a trainable form in Theorem \ref{thm:equivalance}, engaging the deterministic encoder $f_e$ to minimize the reconstruction error and a WS distance between the distribution over sequences of latent codes $[\mathbf{Z}^1, \dots, \mathbf{Z}^M]$ and the optimal $\gamma$ over $[\mathbf{C}^1, \dots, \mathbf{C}^M]$. Additionally, this WS distance is further proven to equivalently decompose into the sum of $M$ WS distances between each $\mathbf{Z^m}$ and $\mathbf{C}^m, m=1,\dots,M$. Interestingly, in Corollary \ref{cor:cluster_latent}, we prove that when minimizing the WS distance between the latent code $\mathbf{Z}^m$ and codeword $\mathbf{C}^m$, the codewords tend to flexibly move to the clustering centroids of the latent representations with a control on the proportion of latent representations associated to a centroid. We argue and empirically demonstrate that using the clustering viewpoint of a WS distance to learn the codewords, we can obtain more \textit{controllable} and \textit{better centroids} than using a simple k-means as in VQ-VAE (cf. Sections~\ref{sec:theoretical_development} and \ref{sec:ablation}).    


Moreover, we leverage the developed theory to propose a practical method called \textit{Vector Quantized Wasserstein Auto-Encoder} (VQ-WAE), which utilizes the WS distance to learn a more controllable codebook, resulting in improved the codebook utilization.
We conduct comprehensive experiments to demonstrate our key contributions by comparing with VQ-VAE \citep{van2017neural} and SQ-VAE \citep{takida22a} (\ie, the recent work that can improve the codebook utilization). The experimental results show that our VQ-WAE can achieve better codebook utilization with higher codebook perplexity, hence leading to lower (compared with VQ-VAE) or comparable (compared with SQ-VAE) reconstruction error, with significantly lower reconstructed Fr\'echlet Inception Distance (FID) score \citep{heusel2017gans}. Generally, a better quantizer in the stage-1 can naturally contribute to stage-2 downstream tasks \citep{yu2021vector,Zheng2022MOvq}. To further demonstrate this, we conduct comprehensive experiments on four benchmark datasets. The experimental results indicate that from the codebooks of our VQ-WAE, we can generate better images with lower FID scores. 

Our contributions in this paper can be summarized:
\begin{itemize}
    \item We are the first work that studies learning discrete representations from the generative viewpoint. 
    Subsequently, 
    we develop rigorous and comprehensive theories that equivalently transform the formulation in the generative viewpoint into another trainable form involving a reconstruction term and a WS distance alignment between the latent representations and learnable codewords.
    \item We harvest our theory development to propose the practical method, namely VQ-WAE, that can learn more controllable codebook for improving the codebook utilization and reconstruct/generate better images with lower FID scores.
\end{itemize}


\section{Vector Quantized Wasserstein Auto-Encoder}

We present the theoretical development of our VQ-WAE framework, which connects the viewpoints of the WS distance, generative models, and deep discrete representation learning in Section \ref{sec:theoretical_development}. It is important to note that our theories are specifically developed for the real setting of discrete representation learning, where a deterministic decoder maps a data example to a sequence of latent codes corresponding to a sequence of codewords. This poses a significant challenge in theory development.
Based on the theoretical development, we devise a practical algorithm for VQ-WAE in Section \ref{sec:proposed_method}. All proofs can be found in Appendix \ref{apd:theory}.


\subsection{Theoretical Development}
\label{sec:theoretical_development}
Given a training set $\mathbb{D}=\left\{ x_{1},...,x_{N}\right\} \subset\mathbb{R}^{n_x}$,
we wish to learn a set of \textit{codewords} $C=\left\{ c_{k}\right\} _{k=1}^{K}\subset\mathbb{R}^{K\times n_z}$
on a latent space $\mathcal{Z}$ and \emph{an} \emph{encoder} to map
each data example to a sequence of $M$ codewords, preserving insightful characteristics
carried in the data. We now endow $M$ discrete distributions: $$\mathbb{P}_{c,\pi^m}= \sum_{k=1}^{K}\pi_{k}^{m}\delta_{c_{k}}, m=1,\dots,M$$
with the Dirac delta function $\delta$ and the weights $\pi^m\in\Delta_{K-1}= \{\alpha\geq \boldsymbol{0}: \Vert \alpha\Vert_1 =1\}$ in the $(K-1)$-simplex.   

We denote $\Gamma = \Gamma(\mathbb{P}_{c,\pi^1},...,\mathbb{P}_{c,\pi^M})$ as the set of all joint distributions over sequences of $M$ codewords, admitting $\mathbb{P}_{c,\pi^{1}}, \dots, \mathbb{P}_{c,\pi^{M}}$ as its marginal distributions. Let also define $\mathcal{\pi} = [\pi^{1},\dots,\pi^{M}]$ as the set of all weights. 



From the generative viewpoint, we propose to learn a decoder function $f_{d}:\mathcal{Z}^{M}\rightarrow\mathcal{X}$
(i.e., mapping from $\mathcal{Z}^{M}$ with the latent space $\mathcal{Z}\subset\mathbb{R}^{n_z}$
to the data space $\mathcal{X}$), the codebook $C$, and the weights
$\pi$, to minimize:
\begin{equation}
\min_{C,\pi}\min_{\gamma \in \Gamma}\min_{f^{d}}\mathcal{W}_{d_{x}}\left(f_{d}\#\gamma,\mathbb{P}_{x}\right),\label{eq:push_forward}
\end{equation}
where $\mathbb{P}_{x}=\frac{1}{N}\sum_{n=1}^{N}\delta_{x_{n}}$ is
the empirical data distribution and $d_{x}$ is a cost metric on the
data space.

We interpret the optimization problem (OP) in Eq. (\ref{eq:push_forward})
as follows. Given discrete distributions $\mathbb{P}_{c,\pi^{1:M}}$, we employ a joint distribution $\gamma \in \Gamma$ as a distribution over sequences of $M$ codewords in $C^M$. We then  use the decoder $f_{d}$ to map the sequences of $M$ codewords in $C^M$
to the data space and consider $\mathcal{W}_{d_{x}}\left(f_{d}\#\gamma,\mathbb{P}_{x}\right)$
as the \emph{codebook-data distortion} w.r.t. $f_{d}$ and $\gamma$.  We subsequently
learn $f_{d}$ to minimize the codebook-data distortion given $\gamma$
and finally adjust the codebook $C$, $\pi$, and $\gamma$ to minimize the optimal
codebook-data distortion. To offer more intuition for the OP in
Eq. (\ref{eq:push_forward}), we introduce the following lemma.


\begin{lemma}
\label{cor:distortion_data}Let $C^{*}=\left\{ c_{k}^{*}\right\} _{k},\pi^{*}$, $\gamma^*$,
and $f_{d}^{*}$ be the optimal solution of the OP in Eq. (\ref{eq:push_forward}).
Assume $K^M<N$, then $C^{*}=\left\{ c_{k}^{*}\right\} _{k},\pi^{*}$,
and $f_{d}^{*}$ are also the optimal solution of the following OP:
\begin{equation}
\min_{f_{d}}\min_{\pi}\min_{\sigma_{1:M}\in\Sigma_{\pi}}\sum_{n=1}^{N}d_{x}\left(x_{n},f_{d}\left([c_{\sigma_m(n)}]_{m=1}^M\right)\right),\label{eq:clus_data}
\end{equation}
where $\Sigma_{\pi}$ is the set of assignment functions $\sigma:\left\{ 1,...,N\right\} \goto\left\{ 1,...,K\right\} $
such that for every $m$ the cardinalities $\left|\sigma_m^{-1}\left(k\right)\right|,k=1,...,K$
are proportional to $\pi^m_{k},k=1,...,K$. Here we denote $\sigma_{m}^{-1}\left(k\right)=\left\{ n\in[N]:\sigma_{m}\left(n\right)=k\right\}$ with $[N]= \{1,2,...,N\}$.
\end{lemma}


Lemma \ref{cor:distortion_data} states that for the optimal solution
$C^{*}=\left\{ c_{k}^{*}\right\} ,\pi^{*}$, $\sigma^*_{1:M}$, and $f_{d}^{*}$ of the
OP in (\ref{eq:push_forward}), each $x_n$ is assigned to the centroid $f_d^*([c_{\sigma^{*}_m}(n)]_{m=1}^M)$ which forms optimal clustering centroids of the optimal clustering
solution minimizing the distortion. 
We establish the following theorem
to engage the OP in (\ref{eq:push_forward}) with the latent space.
\begin{theorem}
\label{thm:reconstruct_form} We can equivalently turn the optimization
problem in (\ref{eq:push_forward}) to
\begin{equation}
\min_{C,\pi, f_d}\min_{\gamma \in \Gamma}\min_{\bar{f}_{e}:\bar{f}_{e}\#\mathbb{P}_{x}=\gamma}\mathbb{E}_{x\sim\mathbb{P}_{x}}\left[d_{x}\left(f_{d}\left(\bar{f}_{e}\left(x\right)\right),x\right)\right],\label{eq:reconstruct_form.}
\end{equation}
where $\bar{f}_{e}$ is a \textbf{deterministic discrete} encoder
mapping data example $x$ directly to a sequence of $M$ codewords in $C^M$.
\end{theorem}

Theorem \ref{thm:reconstruct_form} can be interpreted as follows. First, we learn both the codebook $C$ and the weights $\pi$. Next, we glue the codebook distributions $\mathbb{P}_{c,\pi^m}, m=1,\dots,M$ using the joint distribution $\gamma \in \Gamma$. Subsequently, we  seek a \emph{deterministic discrete} encoder $\bar{f}_{e}$ mapping data example $x$ to sequence of $M$ codewords drawn from $\gamma$, concurring with vector quantization and serving our further derivations. Finally, we minimize the reconstruction error of the sequence of $M$ codewords corresponding to $\bar{f}_{e}(x)$ and $x$.


Additionally, $\bar{f}_{e}$ is a deterministic discrete encoder mapping a
data example $x$ directly to a sequence of codewords. To make it trainable,
we replace $\bar{f}_{e}$ by a continuous encoder $f_{e}:\mathcal{X}\rightarrow\mathcal{Z}^M$ with $f_{e}(x) = [f^m_{e}(x)]_{m=1}^M$ (i.e., each $f_{e}^m:\mathcal{X}\rightarrow\mathcal{Z}$) in the following theorem.

\begin{theorem}
\label{thm:equivalance}
If we seek $f_{d}$ and $f_{e}$ in a family
with infinite capacity (e.g., the family of all measurable functions),
the the two OPs of interest in (\ref{eq:push_forward}) and (\ref{eq:reconstruct_form.}) are equivalent to the following OP
\begin{equation}
\min_{C,\pi}\min_{\gamma \in \Gamma}\min_{f_{d},f_{e}}
\begin{Bmatrix}
\mathbb{E}_{x\sim\mathbb{P}_{x}}\left[d_{x}\left(f_{d}\left(Q_C\left(f_{e}\left(x\right)\right)\right),x\right)\right]
 \\
+\lambda\mathcal{W}_{d_{z}}\left(f_{e}\#\mathbb{P}_{x},\gamma\right)
\end{Bmatrix},
\label{eq:reconstruct_form_continuous}
\end{equation}
where $Q_C\left(f_{e}\left(x\right)\right)=[Q_C(f_{e}^m\left(x\right))]_{m=1}^M$ with $Q_C(f_{e}^m\left(x\right))  = \text{argmin}{}_{c\in C}\rho_{z}\left(f_{e}^m\left(x\right),c\right)$ is a quantization operator which returns the sequence of closest codewords
to $f_{e}^m\left(x\right), m=1,\dots,M$ and the parameter $\lambda>0$. Here we overload the quantization operator for both $f_e(x) \in \mathcal{Z}^M$ and $f_{e}^m(x) \in \mathcal{Z}$. Additionally, given $z =[z^m]_{m=1}^M \in \mathcal{Z}^M, \bar{z} = [\bar{z}^m]_{m=1}^M \in \mathcal{Z}^M$, the distance between them is defined as $$d_{z}\left(z,\bar{z}\right)=\frac{1}{M}\sum_{m=1}^{M}\rho_z\left(z^{m},\bar{z}^{m}\right),$$  where $\rho_z$ is a distance on $\mathcal{Z}$.
\end{theorem}

Particularly, we rigorously prove that the OPs of interest
in (\ref{eq:push_forward}), (\ref{eq:reconstruct_form.}), and (\ref{eq:reconstruct_form_continuous})
are equivalent under some mild conditions in Theorem \ref{thm:equivalance}.
This rationally explains why we could solve the OP in (\ref{eq:reconstruct_form_continuous})
for our final tractable solution. Moreover, the OP in (\ref{eq:reconstruct_form_continuous}) conveys important meaningful interpretations. Specifically, by minimizing $\mathcal{W}_{d_{z}}\left(f_{e}\#\mathbb{P}_{x},\gamma\right)$
w.r.t. $C,\pi$ where $\gamma$ admits $\mathbb{P}_{c,\pi^{1:M}}$ as its marginal distributions, we implicitly minimize $\mathcal{W}_{\rho_z}(f_e^m\# \mathbb{P}_x, \mathbb{P}_{c, \pi^m}), m=1, \dots, M$ due to the fact that the former is an upper-bound of the latter as in Lemma \ref{WS_bounds}. Furthermore, in Lemma \ref{WS_bounds}, we also develop a close form for the WS distance of interest, hinting us a practical method.

\begin{lemma}\label{WS_bounds}
    The Wasserstein distance of interest $\min_\pi\min_{\gamma \in \Gamma} \mathcal{W}_{d_{z}}\left(f_{e}\#\mathbb{P}_{x},\gamma\right)$ is upper-bounded by 
    \begin{align}
\frac{1}{M}\sum_{m=1}^{M}  \mathcal{W}_{\rho_{z}}\left(f_{e}^{m}\#\mathbb{P}_{x},\mathbb{P}_{c,\pi^{m}}\right).\label{eq:ws_bounds}
\end{align}
\end{lemma}


According to Lemma \ref{WS_bounds}, the OP of interest in (\ref{eq:reconstruct_form_continuous}) can be replaced by minimizing its upper-bound as follows 
\begin{equation}
\min_{C,\pi}\min_{f_{d},f_{e}}
\begin{Bmatrix}
\mathbb{E}_{x\sim\mathbb{P}_{x}}\left[d_{x}\left(f_{d}\left(Q_C\left(f_{e}\left(x\right)\right)\right),x\right)\right]
 \\
+\frac{\lambda}{M}\sum_{m=1}^{M} \mathcal{W}_{\rho_{z}}\left(f_{e}^{m}\#\mathbb{P}_{x},\mathbb{P}_{c,\pi^{m}}\right)
\end{Bmatrix}.
\label{eq:reconstruct_form_upper_bound}
\end{equation}

We now interpret the WS term $\mathcal{W}_{\rho_{z}}\left(f_{e}^{m}\#\mathbb{P}_{x},\mathbb{P}_{c,\pi^{m}}\right)$ in Corollary \ref{cor:cluster_latent}.

\begin{corollary}
\label{cor:cluster_latent}Given $m \in [M]$, consider minimizing the term: $\min_{f_{e},C}\mathcal{W}_{\rho_{z}}\left(f^m_{e}\#\mathbb{P}_{x},\mathbb{P}_{c,\pi^m}\right)$
in (\ref{eq:reconstruct_form_continuous}), given $\pi^m$ and assume
$K<N$, its optimal solution $f_{e}^{*m}$ and $C^{*}$are also the
optimal solution of the OP:
\begin{equation}
\min_{f_{e},C}\min_{\sigma\in\Sigma_{\pi}}\sum_{n=1}^{N}\rho_{z}\left(f^m_{e}\left(x_{n}\right),c_{\sigma\left(n\right)}\right),\label{eq:clus_latent}
\end{equation}
where $\Sigma_{\pi}$ is the set of assignment functions $\sigma:\left\{ 1,...,N\right\} \goto\left\{ 1,...,K\right\} $
such that the cardinalities $\left|\sigma^{-1}\left(k\right)\right|,k=1,...,K$
are proportional to $\pi^m_{k},k=1,...,K$.
\end{corollary}

Corollary \ref{cor:cluster_latent}~indicates the aim of minimizing
the second term $\mathcal{W}_{\rho_{z}}\left(f^m_{e}\#\mathbb{P}_{x},\mathbb{P}_{c,\pi^m}\right)$. By which, we adjust
the encoder $f_{e}$ and the codebook $C$ such that the codewords of
$C$ become the clustering centroids of the latent representations
$\left\{ f^m_{e}\left(x_{n}\right)\right\} _{n}$ to minimize the \emph{codebook-latent
distortion}. Additionally, at the optimal solution, the optimal assignment function $\sigma^*$, which indicates how latent representations (or data examples) associated with the clustering centroids (i.e., the codewords) has a valuable property, i.e., \textit{the cardinalities $\left|(\sigma^*)^{-1}\left(k\right)\right|,k=1,...,K$
are proportional to $\pi^m_{k},k=1,...,K$}. 

\textbf{Remark:}  Recall the codebook collapse issue, i.e. most of embedded latent vectors are quantized to just few discrete codewords while the other codewords are rarely
used. Corollary \ref{cor:cluster_latent} give us important properties: \textbf{(1)} \textit{we can control the number of latent representations assigned to each codeword by adjust $\pi^m$, guaranteeing all codewords are utilized}, \textbf{(2)} \textit{codewords become the clustering centroids of the associated latent representations to minimize the codebook-latent distortion}, to develop our VQ-WAE framework. Particularly, we propose adding the regularization terms $D_{KL}(\pi^m, \mathcal{U}_K)$ as the Kullback-Leibler divergence between $\pi^m$ and the uniform distribution $\mathcal{U}_K = [\frac{1}{K}]_K$ to regularize $\pi^m$.

\subsection{Practical Algorithm for 
VQ-WAE}
\label{sec:proposed_method}
We now harvest our theoretical development to propose a practical method named \textit{Vector Quantized Wasserstein Auto-Encoder} (VQ-WAE). Particularly, we combine the objective function in (\ref{eq:reconstruct_form_upper_bound}) with the regularization terms $D_{KL}(\pi^m, \mathcal{U}_K), m=1,\dots,M$ and $\mathcal{U}_K=\left[\frac{1}{K}\right]_{K}$ inspired by Corollary \ref{cor:cluster_latent} to arrive at the following OP:   
\begin{equation}
\min_{C,\pi, f_{d},f_{e}}
\begin{Bmatrix}
\mathbb{E}_{x\sim\mathbb{P}_{x}}\left[d_{x}\left(f_{d}\left(Q_C\left(f_{e}\left(x\right)\right)\right),x\right)\right]   \\
+ \frac{\lambda}{M} \times\sum_{m=1}^{M} \mathcal{W}_{\rho_{z}}\left(f_{e}^{m}\#\mathbb{P}_{x},\mathbb{P}_{c,\pi^{m}}\right) \\
+\lambda_r \sum_{m=1}^{M}D_{KL}(\pi^m, \mathcal{U}_K)
\end{Bmatrix},
\label{eq:KL_regularization_OP}
\end{equation}

where $\lambda, \lambda_r > 0$ are two trade-off parameters.

To learn the weights $\pi^m$,
we parameterize $\pi^m = \pi^m(\beta^m) = \text{softmax}(\beta^m), m=1,\dots,M$ with $\beta^m \in \mathbb{R}^K$. 
Additionally, in order to optimize (\ref{eq:KL_regularization_OP}), we have to deal with $M$ WS distances $\mathcal{W}_{\rho_{z}}\left(f_{e}^{m}\#\mathbb{P}_{x},\mathbb{P}_{c,\pi^{m}}\right)$ with $m=1,...,M$. Therefore, we proposed to use entropic dual form of optimal transport \citep{genevay2016stochastic} which enable us to compute these WS distances in parallel by matrix computation from current deep learning framework.

At each iteration, we sample a mini-batch $x_1,...,x_B$
and then solve the above OP by updating $f_{d},f_{e}$ and $C,\beta^{1..M}$
based on this mini-batch as follows. Let us denote 
$$\mathbb{P}_{B}=\frac{1}{B}\sum_{i=1}^{B}\delta_{x_{i}}$$
as the empirical distribution over the current batch. 

For each mini-batch, we replace $\mathcal{W}_{\rho_{z}}\left(f_{e}^{m}\#\mathbb{P}_{x},\mathbb{P}_{c,\pi^{m}}\right)$ by $\mathcal{W}_{\rho_{z}}\left(f_{e}^{m}\#\mathbb{P}_{B},\mathbb{P}_{c,\pi^{m}}\right)$
and approximate it with
entropic regularized duality form $\mathcal{R}^m_{WS}$ (see Eq. (\ref{apd:entropic_dual_form}) in Appendix \ref{alg_learn_pi_entropic_dual}) as follows:

\begin{equation}
\begin{split}
\mathcal{R}^m_{WS}=\max_{\phi^{m}}
\left\{ 
\frac{1}{B}\sum_{i=1}^{B} 
\left [ -\epsilon\log
\left ( 
\sum_{k=1}^{K}\pi^m_k
\left [ 
\exp 
\Biggl\{
\right. \right. \right.  \right. \Biggr.
\\
\Biggl. \left. \left. \left. \left.   \frac{-\rho_z(f_e^m(x_i),c_k)+\phi^m\left ( c_k\right )}{\epsilon} 
\Biggr\} 
\right ] 
\right ) 
\right ]
+\sum_{k=1}^{K}\pi^m_k\phi^m(c_k)
\right\}
\end{split}
\label{eq:approximation_entropic_form}
\end{equation}
where $\phi^m$ is the Kantorovich potential network.

Substituting (\ref{eq:approximation_entropic_form}) into (\ref{eq:KL_regularization_OP}), we reach final OP to update 
$f_{d},f_{e},C,\left \{ \beta^m \right \}_{m=1}^M$ for each mini-batch:
\begin{equation}
\min_{C,\left \{ \beta^m \right \}_{m=1}^M}\min_{f_{d},f_{e}}
\begin{Bmatrix}
\frac{1}{B}\sum_{i=1}^{B}d_{x}\left(f_{d}\left(Q\left(f_{e}\left(x_{i}\right)\right)\right)\right)   \\
+ \frac{\lambda}{M} \sum_{m=1}^{M}\mathcal{R}^m_{WS}  \\
+ \lambda_r \sum_{m=1}^{M}D_{KL}\left(\pi^{m}\left(\beta^{m}\right), \mathcal{U}_K\right)
\end{Bmatrix}.
\label{eq:final_opt}
\end{equation}

We use the copy gradient trick \citep{van2017neural} to deal with the back-propagation
from decoder to encoder for reconstruction term. 
The pseudocode of our VQ-WAE is summarized in Algorithm~\ref{alg:WQVAE}.

\begin{algorithm}[!]
\caption{VQ-WAE\label{alg:WQVAE}}
\label{alg:ens_optim} \begin{algorithmic}[1] 
\STATE \textbf{Initialize}: encoder $f_e$, decoder $f_d$, codebook $C$ and $\left \{ \pi^m = \text{softmax}(\beta^m), \phi^m \right \}_{m=1}^M$.
\FOR{iter {\bfseries in} batch-iterations}
\STATE Sample a mini-batch of samples $x_{1},...,x_{B}$ forming the empirical batch distribution $\mathbb{P}_B$.
\STATE Encode: $z_{1...B}=f_e(x_{1...B})$ 
\STATE Quantize: $\bar{z}_{1..B}=Q_C(z_{1...B})$ 
\STATE Decode: $\Bar{x}_{1...B}=f_d(\Bar{z}_{1...B})$
\FOR{iter {\bfseries in} $\phi$-iterations}
\STATE Optimize $\left \{ \phi^m \right \}_{m=1}^M$ by maximizing the objective in (\ref{eq:approximation_entropic_form}).
\ENDFOR
\STATE Optimize $f_e$, $f_d$, $\left \{ \beta^m \right \}_{m=1}^M$ and $C$ by minimizing the objective in (\ref{eq:final_opt}).

\ENDFOR
\STATE \textbf{Return:} The optimal $f_e$, $f_d$ and $C$.
\end{algorithmic} 

\end{algorithm}

\section{Related Work}
The Variational Auto-Encoder (VAE) was initially introduced by \citet{kingma2013auto} for learning continuous representations. However, learning discrete latent representations has proven to be much more challenging due to the difficulty of accurately evaluating the gradients required for training the models. To make the gradients tractable, one possible solution is to apply the  Gumbel Softmax reparameterization trick \citep{jang2016categorical} to VAE, which allows us to estimate stochastic gradients for updating the models. Although this technique provides gradients with low variance, it introduces a high-bias gradient estimator. 
Another possible solution is to employ the REINFORCE algorithm \citep{williams1992simple}, which is unbiased but has a high variance. 
Furthermore, these two techniques can be combined in a complementary manner \citep{tucker2017rebar}.

To facilitate the learning of discrete latent codes, VQ-VAE \citep{van2017neural} employs a deterministic encoder/decoder architecture and encourages the codebooks to represent the clustering centroids of the latent representations. Additionally, the copy gradient trick is utilized to back-propagate gradients from the decoder to the encoder \citep{bengio2013estimating}.  Several subsequent works have extended VQ-VAE, notably \citet{roy2018theory, wu2020vector}.Particularly, \citet{roy2018theory} uses the Expectation Maximization (EM) algorithm in the bottleneck stage to train the VQ-VAE for improving the quality of the generated images. However, to maintain the stability of this approach, we need to collect a large number of samples on the latent space. \citet{wu2020vector} imposes noises on the latent codes and uses a Bayesian estimator to optimize the quantizer-based representation. The introduced bottleneck Bayesian estimator outputs the posterior mean of the centroids to the decoder and performs soft quantization of the noisy latent codes which have latent representations preserving the similarity relations of the data space. Recently, \citet{takida22a} extends the standard VAE with stochastic quantization and trainable posterior categorical distribution, showing that the annealing of the stochasticity of the quantization process significantly improves the codebook utilization.

Wasserstein (WS) distance has been widely used in various problems \cite{zhao2021neural, nguyen2021tidot, nguyen2021most, le21a, bui2022a}, especially in generative models \citep{arjovsky17a, NIPS2017_892c3b1c, tolstikhin2017wasserstein, dam2019three}. In their work, \citet{arjovsky17a} utilized a dual form of the WS distance to develop the Wasserstein generative adversarial network (WGAN). Subsequently, \citet{NIPS2017_892c3b1c} introduced the gradient penalty trick to enhance the stability of WGAN. In terms of theory development, mostly related to our work is Wasserstein Auto-Encoder \citep{tolstikhin2017wasserstein}, which focuses on learning continuous latent representations while preserving the characteristics of the input data.

\vspace{0.5mm}
\section{Experiments}
\vspace{1mm}

\paragraph{Datasets:} We empirically evaluate the proposed VQ-WAE in comparison with VQ-VAE \citep{van2017neural} that is the baseline method, VQ-GAN \cite{esser2021taming} and recently proposed SQ-VAE \citep{takida22a} which is the state-of-the-art work of improving the codebook usage, on five different benchmark datasets: CIFAR10 \citep{van2017neural}, MNIST \citep{deng2012mnist}, SVHN \citep{netzer2011reading}, CelebA dataset \citep{liu2015deep, takida22a} and the high-resolution images dataset FFHQ.

\paragraph{Implementation:} For a fair comparison, we utilize the same architectures and hyperparameters for all methods. Additionally, in the primary setting, we use a codeword (discrete latent) dimensionality of $64$ and codebook size $\left | C \right|=512$ for all datasets except FFHQ, which has a codeword dimensionality of $256$ and codebook size $\left | C \right|=1024$, while the hyper-parameters $\{\beta, \tau, \lambda\}$ are specified as presented in the original papers, \ie, $\beta=0.25$ for VQ-VAE and VQ-GAN \citep{esser2021taming}, $\tau=1e^{-5}$ for SQ-VAE and $\lambda=1e^{-3}, \lambda_r=1.0$ for our VQ-WAE. The details of the experimental settings are presented in Appendix \ref{apd:expsetting}.

\subsection{Results on Benchmark Datasets}
\label{sec:benchmark}

\begin{table*}[h!]
    \centering
    \caption{Reconstruction performance ($\downarrow:$ the lower the better and $\uparrow:$ the higher the better).}
    \begin{tabular}{lrcrrrrr}
        \toprule 
Dataset & Model & Latent Size & SSIM $\uparrow$ & PSNR $\uparrow$ & LPIPS $\downarrow$ & rFID $\downarrow$ & Perplexity $\uparrow$
\tabularnewline 
\midrule 
CIFAR10 & VQ-VAE  & 8 $\times$ 8 & 0.70 & 23.14 & 0.35 & 77.3 & 69.8  \tabularnewline
 & SQ-VAE  & 8 $\times$ 8 & \textbf{0.80} & \textbf{26.11} & \textbf{0.23} & 55.4 & 434.8 \tabularnewline
 & VQ-WAE  & 8 $\times$ 8 & \textbf{0.80} & 25.93 & \textbf{0.23} & \textbf{54.3} & \textbf{497.3} \tabularnewline
 \midrule 
         
MNIST & VQ-VAE  & 8 $\times$ 8 & 0.98 & 33.37 & 0.02  & 4.8 & 47.2 \tabularnewline
 & SQ-VAE  & 8 $\times$ 8 & \textbf{0.99} & \textbf{36.25} & \textbf{0.01} & 3.2 & 301.8 \tabularnewline
& VQ-WAE  & 8 $\times$ 8 & \textbf{0.99} & 35.71 & \textbf{0.01} & \textbf{2.33} & \textbf{508.4}
\tabularnewline
 \midrule 
 
SVHN & VQ-VAE & 8 $\times$ 8 & 0.88 & 26.94 & 0.17 & 38.5 &  114.6 \tabularnewline
 & SQ-VAE  & 8 $\times$ 8 & \textbf{0.96} & \textbf{35.37} & \textbf{0.06} & 24.8 & 389.8 \tabularnewline
& VQ-WAE  & 8 $\times$ 8 & \textbf{0.96} & 34.62 & 0.07 & \textbf{23.4} & \textbf{485.1} \tabularnewline
 \midrule 
 
CELEBA & VQ-VAE & 16 $\times$ 16 & 0.82 & 27.48 & 0.19 &  19.4 & 48.9 \tabularnewline
 & SQ-VAE  & 16 $\times$ 16 & \textbf{0.89} & \textbf{31.05} & 0.12 & 14.8 & 427.8 \tabularnewline
 & VQ-WAE & 16 $\times$ 16 & \textbf{0.89} & 30.60 & \textbf{0.11} & \textbf{12.2} & \textbf{503.0}
 \tabularnewline
 
 \midrule 

FFHQ & VQ-GAN & 16 $\times$ 16  & 0.6641 & 22.24 & \textbf{0.1175} & 4.42 & 423  \tabularnewline
& VQ-WAE & 16 $\times$ 16 & \textbf{0.6648} & \textbf{22.45} & 0.1245 & \textbf{4.20} & \textbf{1022} \tabularnewline

        \bottomrule
    \end{tabular}
    \label{tab:fids}
\end{table*}

\paragraph{Quantitative assessment:} In order to quantitatively assess the quality of the reconstructed images, we report the results on most common evaluation metrics, including the pixel-level peak signal-to-noise ratio (PSNR), patch-level structure similarity index (SSIM), feature-level LPIPS \citep{zhang2018unreasonable}, and dataset-level Fr\'echlet Inception Distance (FID) \citep{heusel2017gans}. We report the test-set reconstruction results on four datasets in Table \ref{tab:fids}. With regard to the codebook utilization, we employ perplexity score which is defined as $e^{-\sum_{k=1}^K p_{c_k} \log p_{c_k}}$  where $p_{c_k}= \frac{N_{c_k}}{\sum_{i=1}^K N_{c_i}}$ (i.e., $N_{c_i}$ is the number of latent representations associated with the codeword $c_i$) is the probability of the $i^{th}$ codeword being used. 
Note that by formula, $\text{perplexity}_{\max}=\left | C \right|$ as $P(c)$ becomes to the uniform distribution, which means that all the codewords are utilized equally by the model.


We compare VQ-WAE with VQ-VAE, SQ-VAE and VQ-GAN for image reconstruction in Table~\ref{tab:fids}. All instantiations of our model significantly outperform the baseline VQ-VAE under the same compression ratio, with the same network architecture. While the latest state-of-the-art SQ-VAE or VQ-GAN holds slightly better scores for traditional pixel- and patch-level metrics, our method achieves much better rFID scores which evaluate the image quality at the dataset level. Note that our VQ-WAE significantly improves the perplexity of the learned codebook. This suggests that the proposed method significantly improves the codebook usage, resulting in better reconstruction quality. which is further demonstrated in the following qualitative assessment.

\begin{figure}[!h]
\centering
\includegraphics[width=0.48\textwidth]{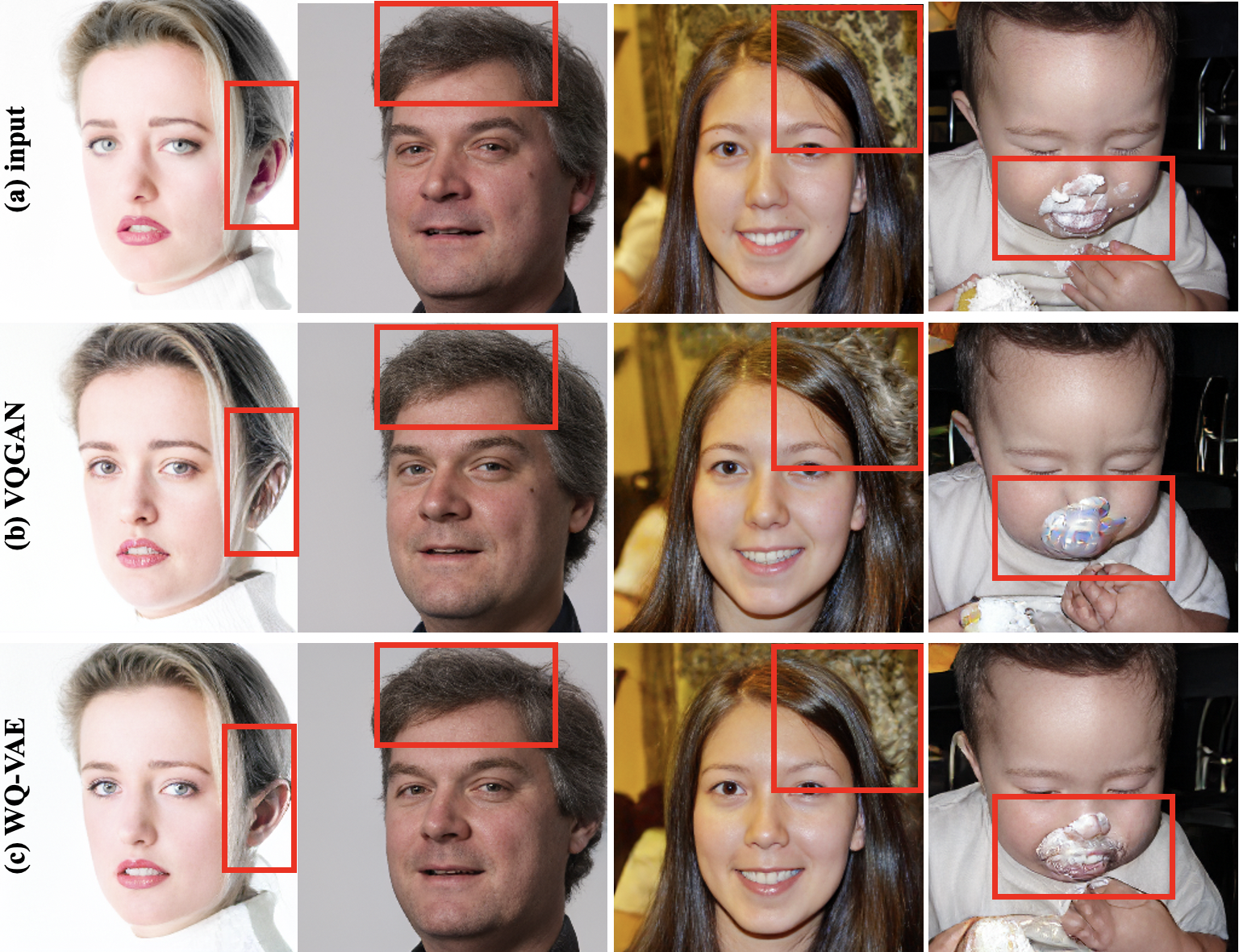}

\caption{Reconstruction results for the FFHQ dataset.}
\label{fig:reconstruction_FFHQ}
\end{figure}

\paragraph{Qualitative assessment:} We present the reconstructed samples from  FFHQ (high-resolution images) for qualitative evaluation. It can be clearly seen that the high-level semantic features of the input image and colors are better preserved with VQ-WAE than the baseline. Particularly, we notice that VQ-GAN often produces repeated artifact patterns in image synthesis (see the hair of man is second column in Figure~\ref{fig:reconstruction_FFHQ}) while VQ-WAE does not. This is because VQ-GAN is lack of diversity in the codebook, which will be further analyzed in Section~\ref{sec-codebook-usage}. 
Consequently, the quantization operator embeds similar patches into the same quantization index and ignores the variance in these patches (e.g., VQ-GAN reconstructs the background in third column of Figure~\ref{fig:reconstruction_FFHQ} as hair of woman).

\subsection{Detailed Analysis} \label{sec:ablation}
We run a number of ablations to analyze the properties of VQ-VAE, SQ-VAE and VQ-WAE, in order to assess if our VQ-WAE can simultaneously achieve (i) \textit{efficient codebook usage}, (ii) \textit{reasonable latent representation}.

\subsubsection{Codebook Usage}
\label{sec-codebook-usage}

\begin{table*}[h!]
    \centering
    \caption{Distortion and Perplexity with different codebook sizes.}
    \begin{tabular}{rl|rrrr|rrrr}
\toprule 
\multicolumn{2}{l|}{Dataset} & \multicolumn{4}{c|}{MNIST} & \multicolumn{4}{c}{CIFAR10} \\
\midrule 
\multicolumn{2}{l|}{$\left | C \right |$} & 64 & 128 & 256 & 512 & 64 & 128 & 256 & 512 \\
 \midrule
 \multirow{2}{*}{VQ-VAE} & Perplexity & 47.8 & 70.3 & 52.0 & 47.2 & 24.3 & 44.9 & 85.1 & 69.8\\
 & rFID & 5.9 &  6.2 & 5.2 & 4.8 & 86.6 & 78.9 & 73.6 & 69.8 \\
 \midrule 
 \multirow{2}{*}{SQ-VAE} & Perplexity & 47.4 & 85.4 & 184.8 & 301.8 & 59.5 & 113.2 & 220.0 & 434.8  \\
 
 & rFID & \textbf{4.7} & 4.3 & 3.5 & 3.2 &\textbf{71.5} & \textbf{66.9} & 62.6 & 55.4  \\
 \midrule 
 \multirow{2}{*}{VQ-WAE} & Perplexity & \textbf{60.1} & \textbf{125.3} & \textbf{245.0} & \textbf{508.4} & \textbf{62.2} & \textbf{121.4} & \textbf{250.9} & \textbf{497.3} \\
 
 & rFID & 5.6 & \textbf{3.9} & \textbf{2.8} & \textbf{2.3} & 73.5 & 68.2 & \textbf{60.5} & \textbf{54.3} \\
        \bottomrule
    \end{tabular}
    \label{tab:different_codebooks}
\end{table*}

\begin{figure*}[!h]
  \centering
  \begin{tabular}[b]{c}
    \includegraphics[width=0.47\textwidth]{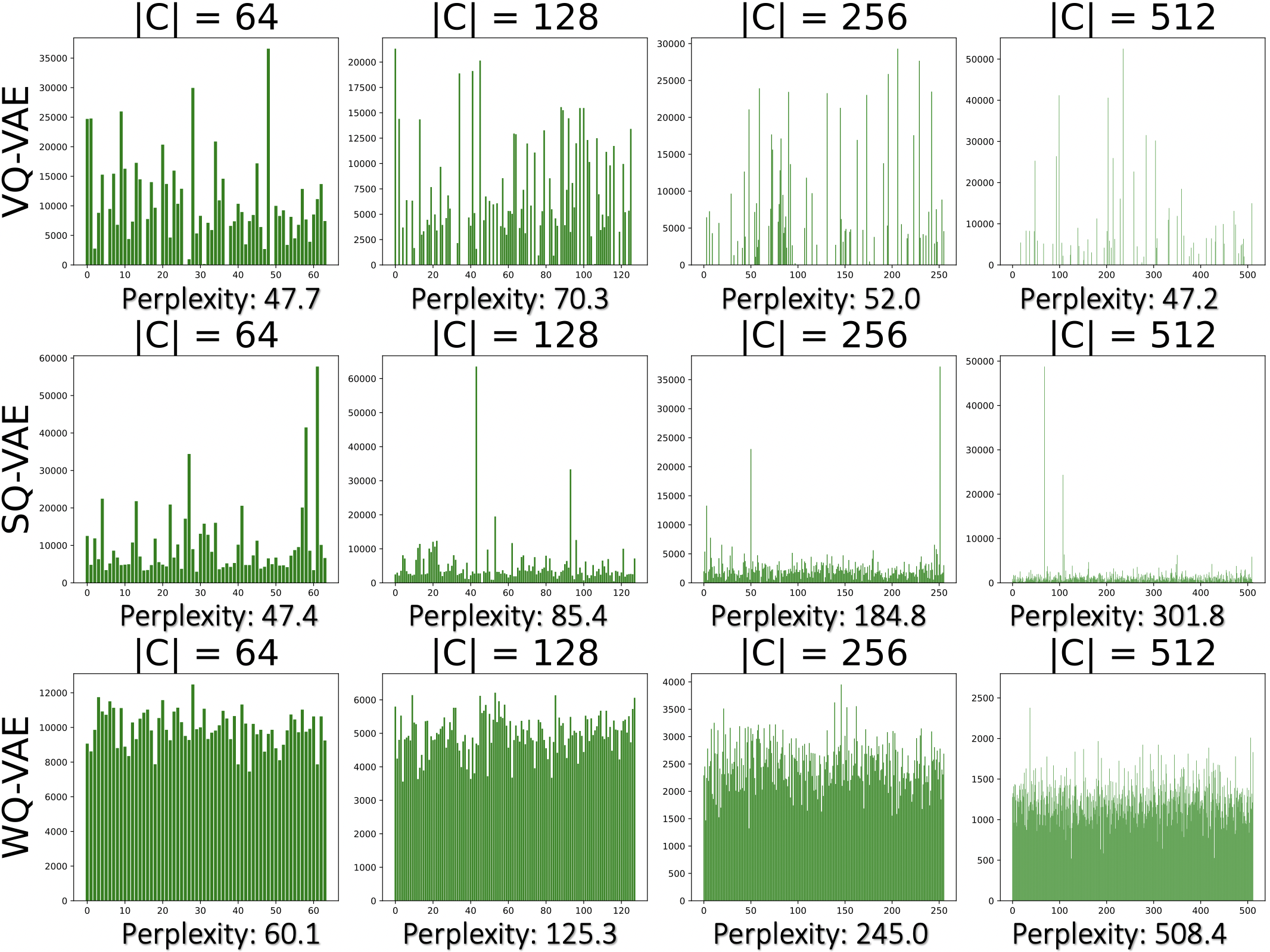} \\
    \small (a) MNIST. \\
    \label{fig:codebook_mnist}
  \end{tabular} 
   \hfill{}
  \begin{tabular}[b]{c}
    \includegraphics[width=0.47\textwidth]{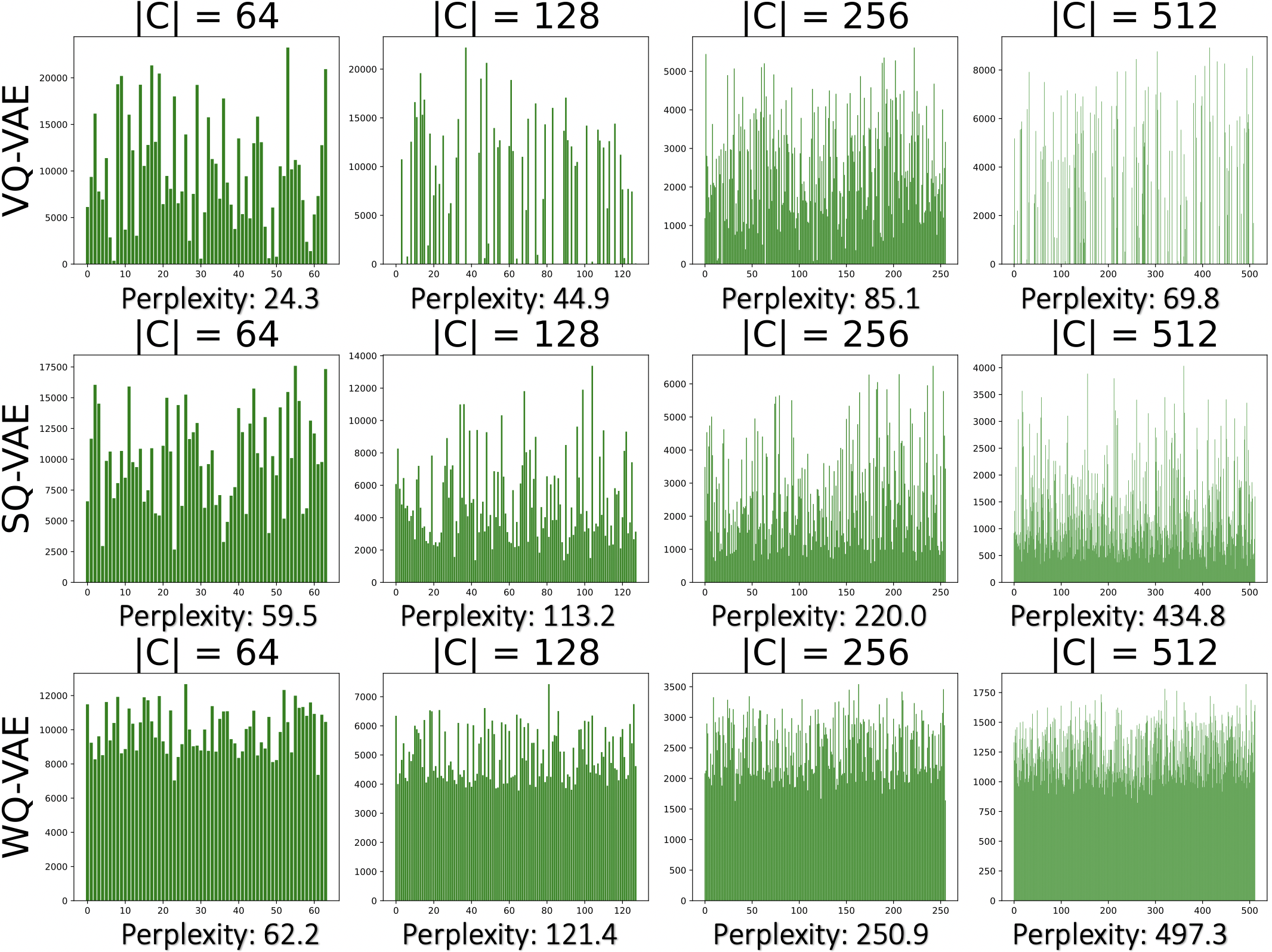} \\
    \small (b) CIFAR10. \\
    \label{fig:codebook_cifar} 
  \end{tabular}
  \vspace{-0mm}
  \caption{Latent distribution over the codebook on test-set.}
  \label{fig:codebook_size_distribution}
\end{figure*}



We observe the codebook utilization of three methods with different codebook sizes $\{64, 128, 256, 512\}$ on MNIST and CIFAR10 datasets. Particularly, we present the reconstruction performance for different settings in Table~\ref{tab:different_codebooks} and the histogram of latent representations over the codebook in Figure~\ref{fig:codebook_size_distribution}. As discussed in Section~\ref{sec:theoretical_development}, the number of used centroids reflects the capability of the latent representations. In other words, it represents the certain amount of information is preserved in the latent space. 

It can be seen from Figure~\ref{fig:codebook_size_distribution} that the latent distribution of VQ-WAE over the codebook is nearly uniform and the codebook's perplexity almost reaches the optimal value (i.e., the value of perplexities reach to corresponding codebook sizes) in different settings. \textit{It is also observed that as the size of the codebook increases, the perplexity of codebook of VQ-WAE also increases, leading to the better reconstruction performance (Table~\ref{tab:different_codebooks}), in line with the analysis in \citep{wu2018variational}}. SQ-VAE also has good codebook utilization as its perplexity is proportional to the size of the codebook. However, it becomes less efficient when the codebook size becomes large, especially in low texture dataset. (i.e., MNIST). On the contrary, the codebook usage of VQ-VAE is less efficient, i.e., there are many zero entries in its codebook usage histogram, indicating that some codewords have never been used (Figure \ref{fig:codebook_size_distribution}). Furthermore, Table~\ref{tab:different_codebooks} also shows the instability of VQ-VAE's reconstruction performance with different codebook sizes.

\subsubsection{controllability of Codebook}  To further underscore the codebook-controllability of VQ-WAE, we proceed to perform the following ablations. Firstly, additional experiments are conducted involving different initializations of $\pi^m$, specifically including Peaked-form (P), Gaussian-form (G), and Uniform-form (U). Our objective is to observe whether the latent distributions over the codebook, obtained after training with a fixed $\pi^m$ configuration, exhibit proportionality to the initial $\pi^m$, thereby effectively demonstrating the controllability. Secondly, we investigate the implications of optimizing $\pi^m$ as opposed to maintaining a fixed state throughout the training process.

\begin{figure}[h!]
\centering
\includegraphics[width=0.48\textwidth]{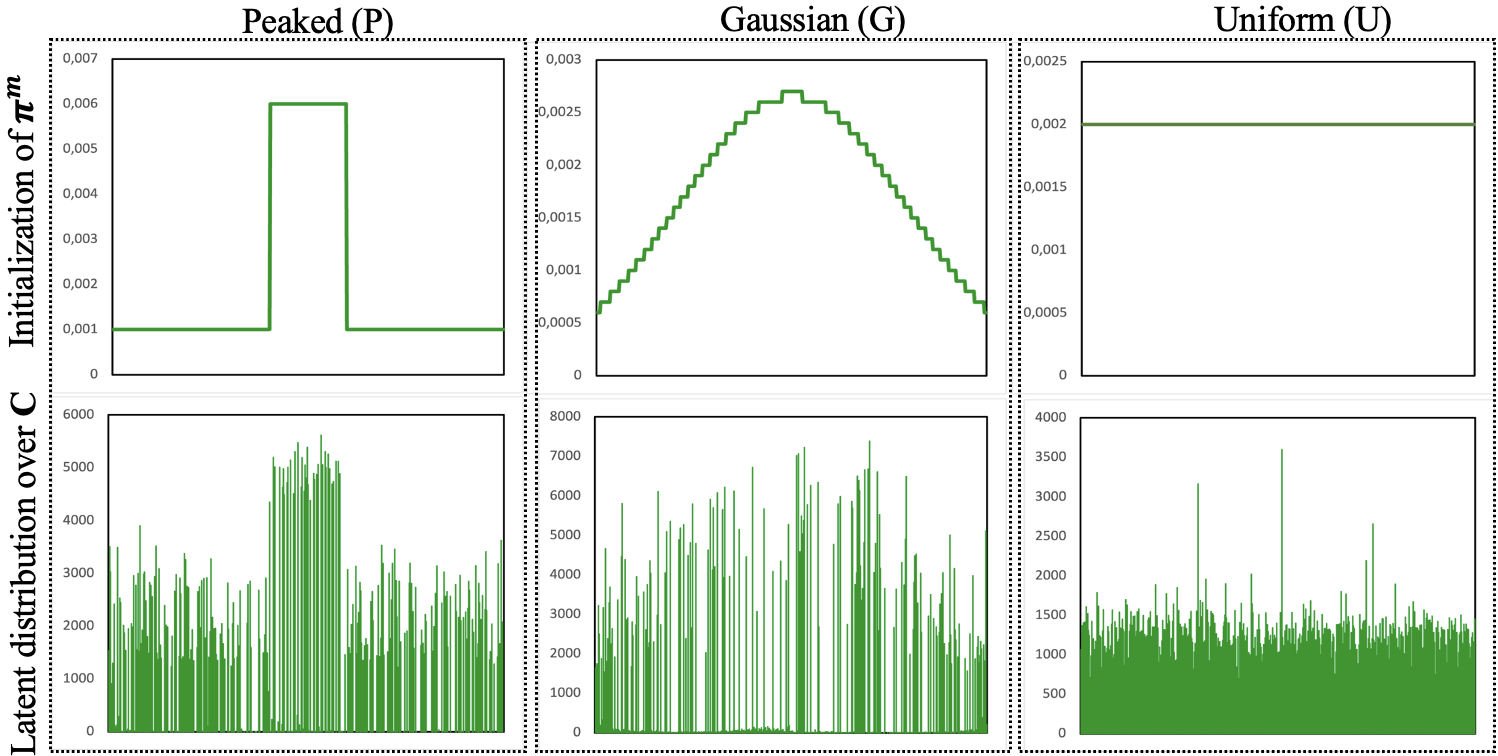}
\vspace{-3mm}
\caption{\textit{Top}. Different initialization of Codebook; \textit{Bottom}. Latent distribution over the codebook $C$ with fixed $\pi^m$.}
\label{fig:init}
\end{figure}

Figure~\ref{fig:init} provides evidence indicating that the latent distributions over the codebook exhibit proportionality to the initial $\pi^m$, thereby serving as a demonstration of the controllability of VQ-WAE's codebook. However, it is important to note that our primary objective is to learn latent representations that accurately approximate the true underlying latent distribution of the data. Consequently, if we have prior knowledge of the true underlying latent distribution of the data, it would be optimal to fix $\pi^m$ accordingly. Nonetheless, in practical scenarios, the true underlying distribution of the data is typically unknown. If the initial $\pi^m$ significantly deviates from the true underlying distribution, it can adversely affect the model's performance. Hence, it is imperative to optimize $\pi^m$ during training process. 

 \begin{table}[h!]
    \centering
    \caption{Reconstruction performance with different codebook initializations (PPL - Perplexity).}
    \label{tab:initialization}
    \begin{tabular}{llccc}
\toprule$\pi^m$ 
& Metric & P & G & U\\
\midrule 
Fixed & rFID & 63.77 & 68.87 & 56.06\\
\midrule 
Fixed & PPL & 229.4 & 165.1 & \textbf{502.6}\\
\midrule 
Updated, $\lambda_r=0.0$ & rFID & 62.04 & 62.16 & 57.49 \\
\midrule 
Updated, $\lambda_r=0.0$ & PPL & 292.5 & 285.6 & 456.5\\
\midrule 
Updated, $\lambda_r=1.0$ & rFID & 60.60 & 60.31 & \textbf{54.30}\\
\midrule 
Updated, $\lambda_r=1.0$ & PPL & 410.0 & 442.8 & 497.3\\
\bottomrule
    \end{tabular}
\end{table}

In such cases, $\pi^m$ will be gradually updated to match the latent distribution. Therefore, our intuition is to initialize $\pi^m$ with a distribution that can easily adapt to arbitrary distributions. The results presented in Table~\ref{tab:initialization} indicate that a uniform initialization is a suitable choice for $\pi^m$. 

It is worth noting that the motivation behind employing KL-regularization is to encourage the utilization of every discrete codeword, thus avoiding the occurrence of certain $\pi^m_k$ values becoming zero (additional discussion regarding the motivation of KL-regularization can be found in Appendix~\ref{sec:KL_analysis}). This feature of VQ-WAE is unique as it allows for the reflection of the latent distribution and enables control over it. Consequently, the Wasserstein distance with KL-regularization in Objective~(\ref{eq:KL_regularization_OP}) serves to match the codebook distribution with the latent data distribution, while also ensuring the utilization of all codewords. This guarantees the robustness of the model.

\subsubsection{Visualization of Latent Representation}

\begin{figure}[h!]
\centering
\includegraphics[width=0.48\textwidth]{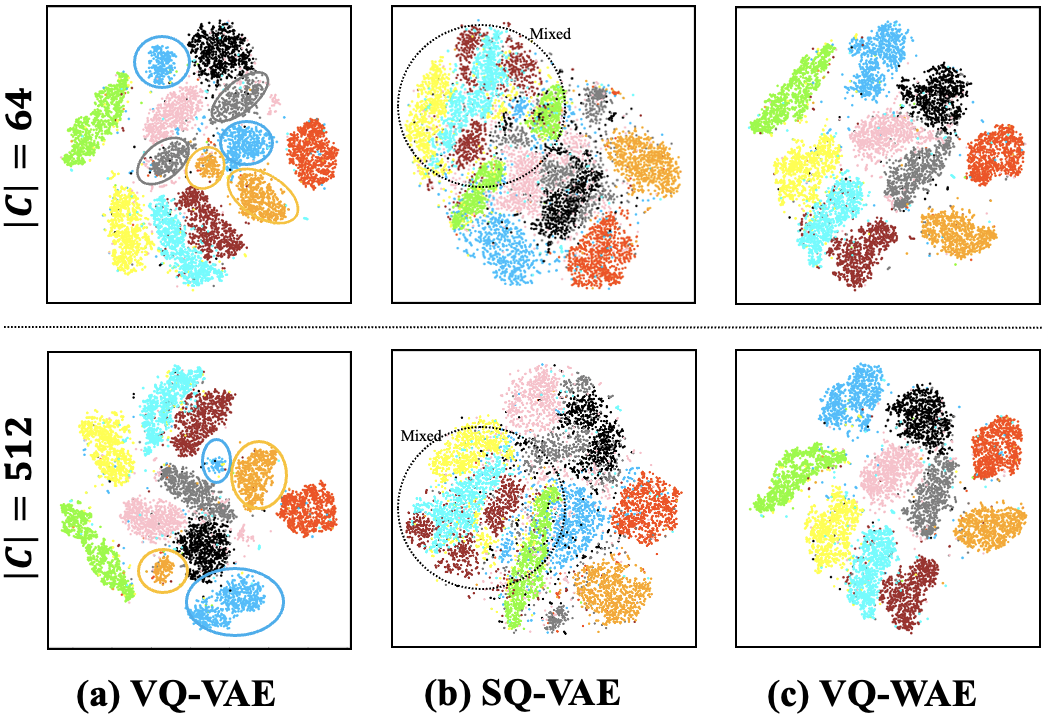}
\vspace{-3mm}
\caption{The t-SNE feature visualization on the MNIST dataset (different colors for different digits).}
\label{fig:tsne}
\vspace{-3mm}
\end{figure}

\paragraph{T-SNE visualization.} To better understand the codebook's representation power, we employ t-SNE~\citep{Maaten08visualizingdata} to visualize the latent that have been learned by VQ-VAE, SQ-VAE and VQ-WAE on the MNIST dataset with two codebook sizes of $64$ and $512$. Figure \ref{fig:tsne} shows the latent distributions of different classes in the latent space, in which the samples are colored accordingly to their class labels. Figure \ref{fig:tsne}c shows that representations from different classes of VQ-WAE are well clustered (i.e., each class focuses on only one cluster) and clearly separated to other classes. In contrast, the representations of some classes in VQ-VAE and SQ-VAE are distributed to several clusters and or mixed to each other (Figure \ref{fig:tsne}a,b). Moreover, the class-clusters of SQ-VAE are uncondensed and tend to overlap with each other.
These results suggest that the representations learned by VQ-WAE can better preserve the similarity relations of the data space better than the other baselines.

\paragraph{Single-layer Classification on latent space.} We train a separate single-layer classifier using the latent representation from auto-encoders (VQ-VAE, SQ-VAE and VQ-WAE) as input. We did not optimize autoencoder's parameters with respect to the classifier’s loss to measure the unsupervised representation learning performance of auto-encoders. 

\begin{table}[h!]
    \centering
    \caption{Single-layer classification accuracy on latent space.}
    \label{tab:classification}
    \begin{tabular}{lccc}
    \toprule
        Dataset & VQ-VAE & SQ-VAE & VQ-WAE\\
         \midrule 
        Cifar10 & 43.21 & 46.17 & \textbf{50.19} \\
         \midrule 
        Mnist & 95.12 & 94.48 & \textbf{95.62}\\
         \midrule 
        SVHN & 35.10 & 36.73 & \textbf{38.38}\\
         \bottomrule
    \end{tabular}
\end{table}

It can be seen from Table~\ref{tab:classification} that VQ-WAE obtained higher performance compared to SQ-VAE, further demonstrating the better quality of a learned representation of VQ-WAE.
\color{black}

\subsubsection{Image Generation}

As discussed in the previous section, VQ-WAE is able to optimally utilize its codebook, leading to meaningful and diverse codewords that naturally improve the image generation. To confirm this ability, we perform the image generation on the benchmark datasets. Since the decoder reconstructs images directly from the discrete embeddings, we only need to model a prior distribution over the discrete latent space (i.e., codebook) to generate images. 
We employ a conventional autoregressive model, the CNN-based PixelCNN \citep{van2016conditional}, to estimate a prior distribution over the discrete latent space of VQ-VAE, SQ-VAE and VQ-WAE on CIFAR10, MNIST, SVHN and CelebA. The details of generation settings are presented in Section 3.2 of the supplementary material. The quantitative results in Table~\ref{tab:generation} indicate that the codebook of VQ-WAE leads to a better generation ability baselines.

\begin{table}[h!]
    \centering
    
    \caption{FID scores of unconditional (U) and class-conditional (C) generated images.}
    \begin{tabular}{lrcrr}
        \toprule 
Dataset & Model & Latent size & U & C
        \tabularnewline 
        \midrule 
CIFAR10 & VQ-VAE & $8 \times 8$ & 117.49 &  117.16 \tabularnewline
 & SQ-VAE & $8 \times 8$ &  103.78 &  90.74 \tabularnewline
& VQ-WAE  & $8 \times 8$ &  \textbf{87.73} &  \textbf{88.51}\tabularnewline
         \midrule 
         
MNIST & VQ-VAE  & $8 \times 8$ &  27.01  & 25.56 \tabularnewline
 & SQ-VAE & $8 \times 8$ &  8.93 & 4.94  \tabularnewline
& VQ-WAE & $8 \times 8$ &  \textbf{8.21} & \textbf{3.88} 
\tabularnewline
 \midrule 
SVHN & VQ-VAE & $8 \times 8$ &   62.13  &  64.24 \tabularnewline
 & SQ-VAE & $8 \times 8$ &  31.26 &  36.41 \tabularnewline
& VQ-WAE & $8 \times 8$ &   \textbf{30.71} & \textbf{34.44} \tabularnewline
 \midrule 
CELEBA & VQ-VAE & $16 \times 16$  &    42.0 & - \tabularnewline
 & SQ-VAE & $16 \times 16$ &  29.5 & -  \tabularnewline
& VQ-WAE & $16 \times 16$ &  \textbf{28.8} & - \tabularnewline

        \bottomrule
    \end{tabular}
    \label{tab:generation}
    \vspace{-3mm}
\end{table}


\color{black}

\section{Conclusion}
In this paper, we study discrete deep representation learning from the generative perspective. By leveraging with the nice properties of the WS distance, we develop rigorous and rich theories to turn the generative-inspired formulation to an equivalent trainable form relevant to a reconstruction term and the WS distances between latent representations and the codeword distributions. We harvest our theory development   to propose Vector Quantized Wasserstein Auto-Encoder (VQ-WAE). We conduct comprehensive experiments to show that our VQ-WAE utilizes the codebooks more efficiently than the baselines, hence leading to better reconstructed and generated image quality. 
Additionally, the ablation study shows our proposed framework can optimally utilize the codebook, resulting diverse codewords, allowing VQ-WAE to produce better reconstructions of data examples and more reasonable geometry of the latent manifold. 

Moreover, the OP in \ref{eq:reconstruct_form.} in Theorem \ref{thm:equivalance} hints us a question about learning the joint distribution $\gamma$ over $\mathbb{P}_{c,\pi^m}, m=1,\dots,M$, which if learned appropriately can be served as a distribution over the sequences of codewords in a generative model. Certainly, we can employ a learnable auto-regressive model to characterize $\gamma$ and train it together with the codewords, encoder, and decoder. Currently, we resort a simple solution by minimizing a relevant upper-bound. We leave the problem of learning $\gamma$ for our future research.




\section*{Acknowledgements}
Dinh Phung and Trung Le gratefully acknowledge the support by the US Airforce FA2386-21-1-4049 grant and the Australian Research Council ARC DP230101176 project.  Trung Le was was further supported by the ECR Seed grant of Faculty of Information Technology, Monash University. 


\nocite{langley00}

\bibliography{conference}
\bibliographystyle{icml2023}

\newpage
\appendix
\onecolumn

\section*{Appendix}
This appendix is organized as follows:
\begin{itemize}
    \item In Section \ref{apd:theory}, we present all proofs for theory developed in the main paper.
    \item In Section \ref{alg_learn_pi_entropic_dual}, we present the detail of practical algorithm for VQ-WAE.
     \item In Section \ref{sec:KL_analysis}, we delve deeper into the motivation behind KL regularization and conduct an analysis of the parameters $\lambda$ and $\lambda_r$.
     
    \item In Section \ref{apd:expsetting}, we present experimental settings and implementation specification of VQ-WAE.
\end{itemize}

\section{Theoretical Development} \label{apd:theory}

\begin{lemma} 

\label{cor:appendix_distortion_data}
\textbf{(Lemma \ref{cor:distortion_data} in the main paper)} 
Let $C^{*}=\left\{ c_{k}^{*}\right\} _{k},\pi^{*}$, $\gamma^*$,
and $f_{d}^{*}$ be the optimal solution of the OP in Eq. (\ref{eq:push_forward}).
Assume $K^M<N$, then $C^{*}=\left\{ c_{k}^{*}\right\} _{k},\pi^{*}$,
and $f_{d}^{*}$ are also the optimal solution of the following OP:
\begin{equation}
\min_{f_{d}}\min_{\pi}\min_{\sigma_{1:M}\in\Sigma_{\pi}}\sum_{n=1}^{N}d_{x}\left(x_{n},f_{d}\left([c_{\sigma_m(n)}]_{m=1}^M\right)\right),\label{eq:clus_data}
\end{equation}
where $\Sigma_{\pi}$ is the set of assignment functions $\sigma:\left\{ 1,...,N\right\} \goto\left\{ 1,...,K\right\} $
such that for every $m$ the cardinalities $\left|\sigma_m^{-1}\left(k\right)\right|,k=1,...,K$
are proportional to $\pi^m_{k},k=1,...,K$. Here we denote $\sigma_{m}^{-1}\left(k\right)=\left\{ n\in[N]:\sigma_{m}\left(n\right)=k\right\}$ with $[N]= \{1,2,...,N\}$.

\end{lemma}

\textbf{Proof of Lemma} \ref{cor:appendix_distortion_data}

$\gamma \in \Gamma$ is a distribution over $C^M$ with $\gamma([c_{i_1},\dots, c_{i_M}])$ satisfying $\sum_{i_{1},..,i_{m-1},i_{m}=k,i_{m+1},...,i_{M}}\gamma\left([c_{i_{1}},..,c_{i_{M}}]\right)=\pi_{k}^{m}$.

$f_d \# \gamma$ is a distribution over $f_d([c_{i_1},\dots,c_{i_M}]$ with the mass $\gamma([c_{i_1},\dots, c_{i_M}])$ or in other words, we have
\[
f_{d}\#\gamma=\sum_{i_1,...,i_M}\gamma([c_{i_1},\dots, c_{i_M}])\delta_{f_d([c_{i_1},\dots,c_{i_M}])}.
\]

Therefore, we reach the following OP:
\begin{equation}
\min_{C,\pi}\min_\gamma\min_{f_{d}}\mathcal{W}_{d_{x}}\left(\frac{1}{N}\sum_{n=1}^{N}\delta_{x_{n}},\sum_{i_1,...,i_M}\gamma([c_{i_1},\dots, c_{i_M}])\delta_{f_d\left([c_{i_1},\dots,c_{i_M}]\right)}\right).\label{eq:appendix_push_forward-1}
\end{equation}

By using the Monge definition, we have 
\begin{align*}
\mathcal{W}_{d_{x}}\left(\frac{1}{N}\sum_{n=1}^{N}\delta_{x_{n}},\sum_{i_1,...,i_M}\gamma([c_{i_1},\dots, c_{i_M}])\delta_{f_d([c_{i_1},\dots,c_{i_M}])}\right) & =\min_{T:T\#\mathbb{P}_{x}=f_{d}\#\gamma}\mathbb{E}_{x\sim\mathbb{P}_{x}}\left[d_{x}\left(x,T\left(x\right)\right)\right]\\
 & =\frac{1}{N}\min_{T:T\#\mathbb{P}_{x}=f_{d}\#\gamma}\sum_{n=1}^{N}d_{x}\left(x_{n},T\left(x_{n}\right)\right).
\end{align*}

Since $T\#\mathbb{P}_{x}=f_{d}\#\gamma$, $T\left(x_{n}\right)=f_{d}\left([c_{i_1},\dots,c_{i_M}]\right)$
for some $i_1,...,i_M$. Additionally, $\left|T^{-1}\left(f_d([c_{i_1},\dots,c_{i_M}])\right)\right|,k=1,...,K$
are proportional to $\gamma([c_{i_1},\dots,c_{i_M}])$. Denote $\sigma_1,...,\sigma_M:\left\{ 1,...,N\right\} \goto\left\{1,\dots,K\right\} $
such that $T\left(x_{n}\right)=f_d([c_{\sigma_1\left(n\right)},\dots, c_{\sigma_M\left(n\right)}]),\forall i=1,...,N$,
we have $\sigma_1,\dots,\sigma_M\in\Sigma_{\pi}$. It follows that 
\[
\mathcal{W}_{d_{x}}\left(\frac{1}{N}\sum_{n=1}^{N}\delta_{x_{n}},\sum_{i_1,...,i_M}\gamma([c_{i_1},\dots, c_{i_M}])\delta_{f_d\left([c_{i_1},\dots,c_{i_M}]\right)}\right)=\frac{1}{N}\min_{\sigma_{1:M}\in\Sigma_{\pi}}\sum_{n=1}^{N}d_{x}\left(x_{n},f_d\left([c_{i_1},\dots,c_{i_M}]\right)\right).
\]

Finally, the the optimal solution of the OP in Eq. (\ref{eq:appendix_push_forward-1})
is equivalent to
\[
\min_{f_{d}}\min_{C,\pi}\min_{\sigma_{1:M}\in\Sigma_{\pi}}\sum_{n=1}^{N}d_{x}\left(x_{n},f_d\left([c_{i_1},\dots,c_{i_M}]\right)\right),
\]
which directly implies the conclusion because we have
\[
|\sigma_{m}^{-1}\left(k\right)| \propto \sum_{i_{1},...,i_{m-1},i_{m}=k,i_{m+1},...,i_{M}}\gamma\left([c_{i_{1}},\dots,c_{i_{M}}]\right)=\pi_{k}^{m}.
\]

\begin{theorem}
\label{thm:appendix_reconstruct_form}\textbf{(Theorem \ref{thm:reconstruct_form} in the main paper)} We can equivalently turn the optimization
problem in (\ref{eq:push_forward}) to
\begin{equation}
\min_{C,\pi, f_d}\min_{\gamma \in \Gamma}\min_{\bar{f}_{e}:\bar{f}_{e}\#\mathbb{P}_{x}=\gamma}\mathbb{E}_{x\sim\mathbb{P}_{x}}\left[d_{x}\left(f_{d}\left(\bar{f}_{e}\left(x\right)\right),x\right)\right],\label{eq:appendix_reconstruct_form.}
\end{equation}
where $\bar{f}_{e}$ is a \textbf{deterministic discrete} encoder
mapping data example $x$ directly to a sequence of $M$ codewords in $C^M$.
\end{theorem}
\textbf{Proof of Theorem \ref{thm:appendix_reconstruct_form}}

We first prove that the OP of interest in (\ref{eq:push_forward})
is equivalent to 
\begin{equation}
\min_{C,\pi,f_{d}}\min_{\gamma \in \Gamma}\min_{\bar{f}_{e}:\bar{f}_{e}\#\mathbb{P}_{x}=\gamma}\mathbb{E}_{x\sim\mathbb{P}_{x},[c_{i_1},\dots,c_{i_M}]\sim\bar{f}_{e}\left(x\right)}\left[d_{x}\left(f_{d}\left([c_{i_1},\dots,c_{i_M}]\right),x\right)\right],\label{eq:appendix_reconstruct_form.-1}
\end{equation}
where $\bar{f}_{e}$ is a \textbf{stochastic discrete} encoder mapping a
data example $x$ directly to sequences of $M$ codewords. To this end, we prove
that
\begin{equation}
\mathcal{W}_{d_{x}}\left(f_{d}\#\gamma,\mathbb{P}_{x}\right)=\min_{\bar{f}_{e}:\bar{f}_{e}\#\mathbb{P}_{x}=\gamma}\mathbb{E}_{x\sim\mathbb{P}_{x},[c_{i_1},\dots,c_{i_M}]\sim\bar{f}_{e}\left(x\right)}\left[d_{x}\left(f_{d}\left([c_{i_1},\dots,c_{i_M}]\right),x\right)\right],\label{eq:appendix_ws_distance_latent}
\end{equation}
where $\bar{f}_{e}$ is a \textbf{stochastic discrete} encoder mapping
data example $x$ directly to the codebooks. 

Let $\bar{f}_{e}$ be a \textbf{stochastic discrete} encoder such
that $\bar{f}_{e}\#\mathbb{P}_{x}=\gamma$ (i.e., $x\sim\mathbb{P}_{x}$
and $[c_{i_1}.\dots,c_{i_M}]\sim\bar{f}_{e}\left(x\right)$ implies $[c_{i_1}.\dots,c_{i_M}]\sim\gamma$).
We consider $\alpha_{d,c}$ as the joint distribution of $\left(x,[c_{i_1}.\dots,c_{i_M}]\right)$
with $x\sim\mathbb{P}_{x}\text{ and } [c_{i_1}.\dots,c_{i_M}]\sim\bar{f}_{e}\left(x\right)$.
We also consider $\alpha_{fc,d}$ as the joint distribution including
$\left(x,x'\right)\sim\alpha_{fc,d}$ where $x\sim\mathbb{P}_{x},$$[c_{i_1}.\dots,c_{i_M}]\sim\bar{f}_{e}\left(x\right)$,
and $x'=f_{d}\left([c_{i_1}.\dots,c_{i_M}]\right)$. This follows that $\alpha_{fc,d}\in\Gamma\left(f_{d}\#\gamma,\mathbb{P}_{x}\right)$
which admits $f_{d}\#\gamma$ and $\mathbb{P}_{x}$ as
its marginal distribution have:
\begin{align*}
\mathbb{E}_{x\sim\mathbb{P}_{x},[c_{i_1}.\dots,c_{i_M}]\sim\bar{f}_{e}\left(x\right)}\left[d_{x}\left(f_{d}\left([c_{i_1}.\dots,c_{i_M}]\right),x\right)\right] & =\mathbb{E}_{(x,[c_{i_1}.\dots,c_{i_M}])\sim\alpha_{d,c}}\left[d_{x}\left(f_{d}\left([c_{i_1}.\dots,c_{i_M}]\right),x\right)\right] \\
&
\overset{(1)}{=}\mathbb{E}_{\left(x,x'\right)\sim\alpha_{fc,d}}\left[d_{x}\left(x,x'\right)\right]\\
 & \geq\min_{\alpha_{fc,d}\in\Gamma\left(f_{d}\#\gamma,\mathbb{P}_{x}\right)}\mathbb{E}_{\left(x,x'\right)\sim\alpha_{fc,d}}\left[d_{x}\left(x,x'\right)\right]\\
 & =\mathcal{W}_{d_{x}}\left(f_{d}\#\alpha,\mathbb{P}_{x}\right).
\end{align*}

Note that we have the equality in (1) due to $\left(id,f_{d}\right)\#\alpha_{d,c}=\alpha_{fc,d}$. 

Therefore, we reach
\[
\min_{\bar{f}_{e}:\bar{f}_{e}\#\mathbb{P}_{x}=\gamma}\mathbb{E}_{x\sim\mathbb{P}_{x},[c_{i_1}.\dots,c_{i_M}]\sim\bar{f}_{e}\left(x\right)}\left[d_{x}\left(f_{d}\left([c_{i_1}.\dots,c_{i_M}]\right),x\right)\right]\geq\mathcal{W}_{d_{x}}\left(f_{d}\#\gamma,\mathbb{P}_{x}\right).
\]

Let $\alpha_{fc,d}\in\Gamma\left(f_{d}\#\gamma,\mathbb{P}_{x}\right)$.
Let $\alpha_{fc,c}\in\Gamma\left(f_{d}\#\gamma,\gamma\right)$
be a deterministic coupling such that $[c_{i_1}.\dots,c_{i_M}]\sim\gamma$ and
$x=f_{d}\left([c_{i_1}.\dots,c_{i_M}]\right)$ imply $\left([c_{i_1}.\dots,c_{i_M}],x\right)\sim\alpha_{c,fc}$. Using
the gluing lemma (see Lemma 5.5 in \cite{santambrogio2015optimal}), there exists a joint distribution $\alpha\in\Gamma\left(\gamma,f_{d}\#\gamma,\mathbb{P}_{x}\right)$
which admits $\alpha_{fc,d}$ and $\alpha_{fc,c}$ as the corresponding
joint distributions. By denoting $\alpha_{d,c}\in\Gamma\left(\mathbb{P}_{x},\gamma\right)$
as the marginal distribution of $\alpha$ over $\mathbb{P}_{x},\gamma$,
we then have
\begin{align*}
\mathbb{E}_{\left(x,x'\right)\sim\alpha_{fc,d}}\left[d_{x}\left(x,x'\right)\right] & =\mathbb{E}_{\left([c_{i_1}.\dots,c_{i_M}],x',x\right)\sim\alpha}\left[d_{x}\left(x,x'\right)\right]=\mathbb{E}_{\left([c_{i_1}.\dots,c_{i_M}],x\right)\sim\alpha_{d,c},x'=f_{d}\left([c_{i_1}.\dots,c_{i_M}]\right)}\left[d_{x}\left(x,x'\right)\right]\\
 & =\mathbb{E}_{\left([c_{i_1}.\dots,c_{i_M}],x\right)\sim\alpha_{d,c}}\left[d_{x}\left(f_{d}\left([c_{i_1}.\dots,c_{i_M}]\right),x\right)\right] \\
 &
 =\mathbb{E}_{x\sim\mathbb{P}_{x},[c_{i_1}.\dots,c_{i_M}]\sim\bar{f}_{e}\left(x\right)}\left[d_{x}\left(f_{d}\left([c_{i_1}.\dots,c_{i_M}]\right),x\right)\right]\\
 & \geq\min_{\bar{f}_{e}:\bar{f}_{e}\#\mathbb{P}_{x}=\gamma}\mathbb{E}_{x\sim\mathbb{P}_{x},[c_{i_1}.\dots,c_{i_M}]\sim\bar{f}_{e}\left(x\right)}\left[d_{x}\left(f_{d}\left([c_{i_1}.\dots,c_{i_M}]\right),x\right)\right],
\end{align*}
where $\bar{f}_e(x) = \alpha_{d,c}(\cdot \mid x)$.

This follows that
\begin{align*}
\mathcal{W}_{d_{x}}\left(f_{d}\#\gamma,\mathbb{P}_{x}\right) & =\min_{\alpha_{fc,d}\in\Gamma\left(f_{d}\#\gamma,\mathbb{P}_{x}\right)}\mathbb{E}_{\left(x,x'\right)\sim\alpha_{fc,d}}\left[d_{x}\left(x,x'\right)\right]\\
 & \geq\min_{\bar{f}_{e}:\bar{f}_{e}\#\mathbb{P}_{x}=\gamma}\mathbb{E}_{x\sim\mathbb{P}_{x},[c_{i_1}.\dots,c_{i_M}]\sim\bar{f}_{e}\left(x\right)}\left[d_{x}\left(f_{d}\left([c_{i_1}.\dots,c_{i_M}]\right),x\right)\right].
\end{align*}

This completes the proof for the equality in Eq. (\ref{eq:appendix_ws_distance_latent}),
which means that the OP of interest in (\ref{eq:push_forward}) is
equivalent to 
\begin{equation}
\min_{C,\pi,f_{d}}\min_{\gamma \in \Gamma}\min_{\bar{f}_{e}:\bar{f}_{e}\#\mathbb{P}_{x}=\gamma}\mathbb{E}_{x\sim\mathbb{P}_{x},[c_{i_1}.\dots,c_{i_M}]\sim\bar{f}_{e}\left(x\right)}\left[d_{x}\left(f_{d}\left([c_{i_1}.\dots,c_{i_M}]\right),x\right)\right].\label{eq:appendix_reconstruct_form.-1-1}
\end{equation}

We now further prove the above OP is equivalent to
\begin{equation}
\min_{C,\pi,f_{d}}\min_{\gamma \in \Gamma}\min_{\bar{f}_{e}:\bar{f}_{e}\#\mathbb{P}_{x}=\gamma}\mathbb{E}_{x\sim\mathbb{P}_{x}}\left[d_{x}\left(f_{d}\left(\bar{f}_{e}\left(x\right)\right),x\right)\right],\label{eq:appendix_reconstruct_form.-2}
\end{equation}
where $\bar{f}_{e}$ is a \textbf{deterministic discrete} encoder
mapping data example $x$ directly to the codebooks.

It is obvious that the OP in (\ref{eq:appendix_reconstruct_form.-2}) is special
case of that in (\ref{eq:appendix_reconstruct_form.-1-1}) when we limit to
search for deterministic discrete encoders. Given the optimal solution
$C^{*1},\pi^{*1}, \gamma^{*1}, f_{d}^{*1}$, and $\bar{f}_{e}^{*1}$ of the OP in
(\ref{eq:appendix_reconstruct_form.-1-1}), we show how to construct the optimal
solution for the OP in (\ref{eq:appendix_reconstruct_form.-2}). Let us construct
$C^{*2}=C^{*1}$, $f_{d}^{*2}=f_{d}^{*1}$. Given $x\sim\mathbb{P}_{x}$,
let us denote $\bar{f}_{e}^{*2}\left(x\right)=\text{argmin}{}_{[c_{i_1}.\dots,c_{i_M}]}d_{x}\left(f_{d}^{*2}\left([c_{i_1}.\dots,c_{i_M}]\right),x\right)$.
Thus, $\bar{f}_{e}^{*2}$ is a deterministic discrete encoder mapping
data example $x$ directly to a sequence of codewords. We define $\pi_{k}^{*m2}=Pr\left(\bar{f}_{e,m}^{*2}\left(x\right)=c_{k}:x\sim\mathbb{P}_{x}\right),k=1,...,K$ where $\bar{f}_{e}^{*2}\left(x\right)=[\bar{f}_{e,m}^{*2}\left(x\right)]_{m=1}^M$,
meaning that $\bar{f}_{e}^{*2}\#\mathbb{P}_{x}=\gamma^{*2}$, admitting $\mathbb{P}_{c^{*2},\pi^{*m2}},m=1,\dots,M$ as its marginal distributions.
From the construction of $\bar{f}_{e}^{*2}$, we have 
\[
\mathbb{E}_{x\sim\mathbb{P}_{x}}\left[d_{x}\left(f_{d}^{*2}\left(\bar{f}_{e}^{*2}\left(x\right)\right),x\right)\right]\leq\mathbb{E}_{x\sim\mathbb{P}_{x},[c_{i_1}.\dots,c_{i_M}]\sim\bar{f}_{e}^{*1}\left(x\right)}\left[d_{x}\left(f_{d}^{*1}\left([c_{i_1}.\dots,c_{i_M}]\right),x\right)\right].
\]

Furthermore, because $C^{*2},\pi^{*2},f_{d}^{*2},\text{and}\bar{f}_{e}^{*2}$
are also a feasible solution of the OP in (\ref{eq:appendix_reconstruct_form.-2}),
we have
\[
\mathbb{E}_{x\sim\mathbb{P}_{x}}\left[d_{x}\left(f_{d}^{*2}\left(\bar{f}_{e}^{*2}\left(x\right)\right),x\right)\right]\geq\mathbb{E}_{x\sim\mathbb{P}_{x},[c_{i_1}.\dots,c_{i_M}]\sim\bar{f}_{e}^{*1}\left(x\right)}\left[d_{x}\left(f_{d}^{*1}\left([c_{i_1}.\dots,c_{i_M}]\right),x\right)\right].
\]

This means that 
\[
\mathbb{E}_{x\sim\mathbb{P}_{x}}\left[d_{x}\left(f_{d}^{*2}\left(\bar{f}_{e}^{*2}\left(x\right)\right),x\right)\right]=\mathbb{E}_{x\sim\mathbb{P}_{x},[c_{i_1}.\dots,c_{i_M}]\sim\bar{f}_{e}^{*1}\left(x\right)}\left[d_{x}\left(f_{d}^{*1}\left([c_{i_1}.\dots,c_{i_M}]\right),x\right)\right],
\]
and $C^{*2},\pi^{*2}, \gamma^{*2}, f_{d}^{*2},\text{and}\bar{f}_{e}^{*2}$ are also
the optimal solution of the OP in (\ref{eq:appendix_reconstruct_form.-2}).


We now propose and prove the following lemma that is necessary for the proof of Theorem \ref{thm:appendix_equivalance}.
\begin{lemma}
\label{measure_lemma}
Consider $C, \pi, f_d$, and $f_e$ as a feasible solution of the OP in (\ref{eq:reconstruct_form_continuous}). Let us denote $\bar{f}_e^m(x) = argmin_{c} \rho_z(f_e^m(x)),c) = Q_C(x)$, then $\bar{f}_e^m(x)$ is a Borel measurable function and hence also $\Bar{f}_e(x) = [\bar{f}_e^m(x)]_{m=1}^M$ 
\end{lemma}
\textbf{Proof of Lemma \ref{measure_lemma}}.

We denote the set $A_k$ on the latent space as 
\begin{equation*}
A_k = \{z: \rho_z(z,c_k) < \rho_z(z, c_j), \forall j\neq k \} = \{z: Q_C(z) = c_k\}.    
\end{equation*}
$A_k$ is known as  a Voronoi cell w.r.t. the metric $\rho_z$. If we consider a continuous metric $\rho_z$, $A_k$ is a measurable set. Given a Borel measurable function $B$, we prove that $(\bar{f}_e^{m})^{-1}(B)$ is a Borel measurable set on the data space.

Let $B\cap\{c_{1},..,c_{K}\}=\{c_{i_{1}},...,c_{i_{t}}\}$, we prove that $(\bar{f}_e^{m})^{-1}\left(B\right)=\cup_{j=1}^{t}(\bar{f}_e^{m})^{-1}\left(A_{i_{j}}\right)$. Indeed, take $x \in (\bar{f}_e^{m})^{-1}\left(B\right)$, then $(\bar{f}_e^{m})^{-1}(x) \in B$, implying that $(\bar{f}_e^{m})^{-1}(x) = Q_C(x) = c_{i_j}$ for some $j=1,...,t$. This means that $f_e^m(x) \in A_{i_j}$ for some $j=1,...,t$. Therefore, we reach $(\bar{f}_e^{m})^{-1}\left(B\right) \subset\cup_{j=1}^{t}(f_{e}^m)^{-1}\left(A_{i_{j}}\right)$.

We now take $x \in \cup_{j=1}^{t}(f_{e}^m)^{-1}\left(A_{i_{j}}\right)$. Then $f_e^m(x) \in A_{i_j}$ for $j=1,...,t$, hence $\bar{f}_{e}^m(x) = Q_C(x) = c_{i_j}$ for some $j=1,...,t$. Thus, $\bar{f}_{e}^m(x) \subset B$ or equivalently $x \in (\bar{f}_{e}^m)^{-1}\left(B\right)$, implying $(\bar{f}_{e}^m)^{-1}\left(B\right) \supset\cup_{j=1}^{t}(f_{e}^m)^{-1}\left(A_{i_{j}}\right)$.

Finally, we reach $(\bar{f}_{e}^m)^{-1}\left(B\right)=\cup_{j=1}^{t}(f_{e}^m)^{-1}\left(A_{i_{j}}\right)$, which concludes our proof because $f_e^m$ is a measurable function and $A_{i_j}$ are measurable sets. 

\begin{theorem}
\label{thm:appendix_equivalance}\textbf{(Theorem \ref{thm:equivalance} in the main paper)} 
If we seek $f_{d}$ and $f_{e}$ in a family
with infinite capacity (e.g., the family of all measurable functions),
the the two OPs of interest in (\ref{eq:push_forward}) and (\ref{eq:reconstruct_form.}) are equivalent to the following OP
\begin{equation}
\min_{C,\pi}\min_{\gamma \in \Gamma}\min_{f_{d},f_{e}}
\begin{Bmatrix}
\mathbb{E}_{x\sim\mathbb{P}_{x}}\left[d_{x}\left(f_{d}\left(Q_C\left(f_{e}\left(x\right)\right)\right),x\right)\right]
 \\
+\lambda\mathcal{W}_{d_{z}}\left(f_{e}\#\mathbb{P}_{x},\gamma\right),
\end{Bmatrix}
\label{eq:appendix:reconstruct_form_continuous}
\end{equation}
where $Q_C\left(f_{e}\left(x\right)\right)=[Q_C(f_{e}^m\left(x\right))]_{m=1}^M$ with $Q_C(f_{e}^m\left(x\right))  = \text{argmin}{}_{c\in C}\rho_{z}\left(f_{e}^m\left(x\right),c\right)$ is a \emph{quantization operator} which returns the sequence of closest codewords
to $f_{e}^m\left(x\right), m=1,\dots,M$ and the parameter $\lambda>0$. Here we overload the quantization operator for both $f_e(x) \in \mathcal{Z}^M$ and $f_{e}^m(x) \in \mathcal{Z}$. Additionally, given $z =[z^m]_{m=1}^M \in \mathcal{Z}^M, \bar{z} = [\bar{z}^m]_{m=1}^M \in \mathcal{Z}^M$, the distance between them is defined as $d_{z}\left(z,\bar{z}\right)=\frac{1}{M}\sum_{m=1}^{M}\rho_z\left(z^{m},\bar{z}^{m}\right)$ where $\rho_z$ is a distance on $\mathcal{Z}$.
\end{theorem}
\textbf{Proof of Theorem \ref{thm:appendix_equivalance}}.

Given the optimal solution $C^{*1},\pi^{*1},f_{d}^{*1}, \gamma^{*1}$, and $f_{e}^{*1}$
of the OP in (\ref{eq:reconstruct_form_continuous}), we conduct the
optimal solution for the OP in (\ref{eq:reconstruct_form.}). Let
us conduct $C^{*2}=C^{*1},f_{d}^{*2}=f_{d}^{*1}$. We next define
$\bar{f}_{e}^{*2}\left(x\right)=Q_{C^{*1}}\left(f_{e}^{*1}\left(x\right)\right) = Q_{C^{*2}}\left(f_{e}^{*1}\left(x\right)\right)$. We prove that $C^{*2}, \pi^{*2}, f_d^{*2}$, and $\bar{f}_e^{*2}$ are optimal solution of the OP in (\ref{eq:reconstruct_form.}). Define $\gamma^{*2} = Q_{C^{*2}}\# (f_e^{*1}\#\mathbb{P}_x)$. By this definition, we yield $\bar{f}_{e}^{*2}\#\mathbb{P}_{x}=\gamma^{*2}$
and hence $\mathcal{W}_{d_{z}}\left(\bar{f}_{e}^{*2}\#\mathbb{P}_{x},\gamma^{*2}\right)=0$. Therefore, we need to verify the following:

(i) $\bar{f}_e^{*2}$ is a Borel-measurable function.

(ii) Given a feasible solution $C, \pi, f_d, \gamma$, and $\bar{f}_e$ of (\ref{eq:reconstruct_form.}), we have 
\begin{align}
\mathbb{E}_{x\sim\mathbb{P}_{x}}\left[d_{x}\left(f_{d}^{*2}\left(\bar{f}_{e}^{*2}\left(x\right)\right),x\right)\right] & \leq\mathbb{E}_{x\sim\mathbb{P}_{x}}\left[d_{x}\left(f_{d}\left(\bar{f}_{e}\left(x\right)\right),x\right)\right]. \label{eq:ii}
\end{align}
We first prove (i). It is a direct conclusion because the application of Lemma \ref{measure_lemma} to $C^{*1},\pi^{*1},f_{d}^{*1}$, and $f_{e}^{*1}$. 

We next prove (ii). We further derive as 
\begin{align}
&\mathbb{E}_{x\sim\mathbb{P}_{x}}\left[d_{x}\left(f_{d}^{*2}\left(\bar{f}_{e}^{*2}\left(x\right)\right),x\right)\right] +\lambda\mathcal{W}_{d_{z}}\left(\bar{f}_{e}^{*2}\#\mathbb{P}_{x},\gamma^{*2}\right)
\nonumber\\
& = \mathbb{E}_{x\sim\mathbb{P}_{x}}\left[d_{x}\left(f_{d}^{*2}\left(\bar{f}_{e}^{*2}\left(x\right)\right),x\right)\right]
\nonumber\\
& = \mathbb{E}_{x\sim\mathbb{P}_{x}}\left[d_{x}\left(f_{d}^{*1}\left(Q_{C^{*2}}\left(f_{e}^{*1}\left(x\right)\right)\right),x\right)\right]\nonumber\\
& = \mathbb{E}_{x\sim\mathbb{P}_{x}}\left[d_{x}\left(f_{d}^{*1}\left(Q_{C^{*1}}\left(f_{e}^{*1}\left(x\right)\right)\right),x\right)\right]
\nonumber\\
& \leq \mathbb{E}_{x\sim\mathbb{P}_{x}}\left[d_{x}\left(f_{d}^{*1}\left(Q_{C^{*1}}\left(f_{e}^{*1}\left(x\right)\right)\right),x\right)\right]+\lambda\mathcal{W}_{d_{z}}\left(f_{e}^{*1}\#\mathbb{P}_{x},\gamma^{*1}\right).
\label{eq:1}
\end{align}

Moreover,  because $\bar{f}_e\#\mathbb{P}_x = \gamma$ which is a discrete distribution over $C^M$, we obtain $Q_C(\bar{f}_e(x)) = \bar{f}_e(x)$. Note that $C, \pi, f_d$, and $\bar{f}_e$ is also a feasible solution of (\ref{eq:reconstruct_form_continuous}) because $\bar{f}_e$ is also a specific encoder mapping from the data space to the latent space, we achieve  

\begin{align*}
& \mathbb{E}_{x\sim\mathbb{P}_{x}}\left[d_{x}\left(f_{d}\left(Q_{C}\left(\bar{f}_{e}\left(x\right)\right)\right),x\right)\right]+\lambda\mathcal{W}_{d_{z}}\left(\bar{f}_{e}\#\mathbb{P}_{x},\gamma\right)\nonumber\\
& \geq\mathbb{E}_{x\sim\mathbb{P}_{x}}\left[d_{x}\left(f_{d}^{*1}\left(Q_{C^{*1}}\left(\bar{f}_{e}^{*1}\left(x\right)\right),x\right)\right)\right]+\lambda\mathcal{W}_{d_{z}}\left(\bar{f}_{e}^{*1}\#\mathbb{P}_{x},\gamma^{*1}\right).
\end{align*}

Noting that $\bar{f}_e\#\mathbb{P}_x = \gamma$ and $Q_C(\bar{f}_e(x)) = \bar{f}_e(x)$, we arrive at
\begin{align}
& \mathbb{E}_{x\sim\mathbb{P}_{x}}\left[d_{x}\left(f_{d}\left(\bar{f}_{e}\left(x\right)\right),x\right)\right]
\nonumber\\
& \geq\mathbb{E}_{x\sim\mathbb{P}_{x}}\left[d_{x}\left(f_{d}^{*1}\left(Q_{C^{*1}}\left(\bar{f}_{e}^{*1}\left(x\right)\right)\right),x\right)\right]+\lambda\mathcal{W}_{d_{z}}\left(\bar{f}_{e}^{*1}\#\mathbb{P}_{x},\gamma^{*1}\right). \label{eq:2}
\end{align}
Combining the inequalities in (\ref{eq:1}) and (\ref{eq:2}), we obtain Inequality (\ref{eq:ii}) as

\begin{align}
\mathbb{E}_{x\sim\mathbb{P}_{x}}\left[d_{x}\left(f_{d}^{*2}\left(\bar{f}_{e}^{*2}\left(x\right)\right),x\right)\right]
& \leq\mathbb{E}_{x\sim\mathbb{P}_{x}}\left[d_{x}\left(f_{d}\left(\bar{f}_{e}\left(x\right)\right),x\right)\right]. 
\end{align}
This concludes our proof.

\begin{lemma}\label{appendix_WS_bounds}
    The WS of interest $\min_\pi\min_{\gamma \in \Gamma} \mathcal{W}_{d_{z}}\left(f_{e}\#\mathbb{P}_{x},\gamma\right)$ is upper-bounded by
    \begin{align}
\frac{1}{M}\sum_{m=1}^{M}  \mathcal{W}_{\rho_{z}}\left(f_{e}^{m}\#\mathbb{P}_{x},\mathbb{P}_{c,\pi^{m}}\right).\label{eq:appendix_ws_bounds}
\end{align}
\end{lemma}

\textbf{Proof of Lemma \ref{appendix_WS_bounds}}

Let $\alpha^{*m}\in\Gamma\left(f_{e}^{m}\#\mathbb{P}_{x},\mathbb{P}_{c,\pi^{m}}\right)$ be the optimal coupling for the WS distance $\mathcal{W}_{\rho_{z}}\left(f_{e}^{m}\#\mathbb{P}_{x},\mathbb{P}_{c,\pi^{m}}\right)$. We construct a coupling $\alpha\in\Gamma(f_{e}\#\mathbb{P}_{x},\gamma)$ as follows. We first sample $X \sim \mathbb{P}_x$. We then simultaneously sample $C_m \sim \alpha^{*m}(\cdot \mid f_e^m(X)), m=1,\dots, M$. Let $\gamma^*$ be the law of $[C_1,\dots, C_M]$ and $\alpha^*$ be the law of $\left(f_e(X), [C_1,\dots, C_M]\right)$. Let define $\pi^{*m}$ such that $\mathbb{P}_{c,\pi^{*m}}$ is the marginal distribution of $\gamma^*$ over $C_m$. We then have $\gamma^* \in \Gamma(\mathbb{P}_{c,\pi^1},\dots, \mathbb{P}_{c,\pi^M})$ and $\alpha^* \in \Gamma(f_e\#\mathbb{P}_x, \gamma^*)$. It follows that
\begin{align*}
\mathcal{W}_{d_{z}}\left(f_{e}\#\mathbb{P}_{x},\gamma^{*}\right) & =\mathbb{E}_{(Z,[C_{1},...,C_{M}])\sim\alpha^{*}}\left[d_{z}\left(Z,[C_{1},...,C_{M}]\right)\right]\\
= & \mathbb{E}_{(f_{e}\left(X\right),[C_{1},...,C_{M}])\sim\alpha^{*}}\left[d_{z}\left([f_{e}^{1}\left(X\right),\dots,f_{e}^{M}\left(X\right)],[C_{1},...,C_{M}]\right)\right]\\
= & \frac{1}{M}\sum_{m=1}^{M}\mathbb{E}_{(f_{e}^{m}\left(X\right),C_{m})\sim\alpha^{*m}}\left[\rho_{z}\left(f_{e}^{m}\left(X\right),C_{m}\right)\right]\\
= & \frac{1}{M}\sum_{m=1}^{M}\mathcal{W}_{\rho_{z}}\left(f_{e}^{m}\#\mathbb{P}_{x},\mathbb{P}_{c,\pi^{m}}\right).
\end{align*}

\begin{equation}
\min_{\pi}\min_{\gamma\in\Gamma}\mathcal{W}_{d_{z}}\left(f_{e}\#\mathbb{P}_{x},\gamma\right)\leq\mathcal{W}_{d_{z}}\left(f_{e}\#\mathbb{P}_{x},\gamma^{*}\right)=\frac{1}{M}\sum_{m=1}^{M}\mathcal{W}_{\rho_{z}}\left(f_{e}^{m}\#\mathbb{P}_{x},\mathbb{P}_{c,\pi^{m}}\right).\label{eq:leq}
\end{equation}


\begin{corollary}
\label{cor:appendix_cluster_latent}\textbf{(Corollary \ref{cor:cluster_latent} in the main paper)} 
Given $m \in [M]$, consider minimizing the term: $\min_{f_{e},C}\mathcal{W}_{\rho_{z}}\left(f^m_{e}\#\mathbb{P}_{x},\mathbb{P}_{c,\pi^m}\right)$
in (\ref{eq:reconstruct_form_continuous}), given $\pi^m$ and assume
$K<N$, its optimal solution $f_{e}^{*m}$ and $C^{*}$are also the
optimal solution of the OP:
\begin{equation}
\min_{f_{e},C}\min_{\sigma\in\Sigma_{\pi}}\sum_{n=1}^{N}\rho_{z}\left(f^m_{e}\left(x_{n}\right),c_{\sigma\left(n\right)}\right),\label{eq:clus_latent}
\end{equation}
where $\Sigma_{\pi}$ is the set of assignment functions $\sigma:\left\{ 1,...,N\right\} \goto\left\{ 1,...,K\right\} $
such that the cardinalities $\left|\sigma^{-1}\left(k\right)\right|,k=1,...,K$
are proportional to $\pi^m_{k},k=1,...,K$.
\end{corollary}

\textbf{Proof of Corollary \ref{cor:appendix_cluster_latent}}.

By the Monge definition, we have
\begin{align*}
\mathcal{W}_{\rho_{z}}\left(f_{e}^m\#\mathbb{P}_{x},\mathbb{P}_{c,\pi^m}\right) & =\mathcal{W}_{\rho_{z}}\left(\frac{1}{N}\sum_{n=1}^{N}\delta_{f_{e}^m\left(x_{n}\right)},\sum_{k=1}^{K}\pi_{k}^m\delta_{c_{k}}\right)=\min_{T:T\#\left(f_{e}^m\#\mathbb{P}_{x}\right)=\mathbb{P}_{c,\pi^m}}\mathbb{E}_{z\sim f_{e}^m\#\mathbb{P}_{x}}\left[\rho_{z}\left(z,T\left(z\right)\right)\right]\\
= & \frac{1}{N}\min_{T:T\#\left(f_{e}^m\#\mathbb{P}_{x}\right)=\mathbb{P}_{c,\pi^m}}\sum_{n=1}^{N}\rho_{z}\left(f_{e}^m\left(x_{n}\right),T\left(f_{e}^m\left(x_{n}\right)\right)\right).
\end{align*}

Since $T\#\left(f_{e}^m\#\mathbb{P}_{x}\right)=\mathbb{P}_{c,\pi^m}$,
$T\left(f_{e}\left(x_{n}\right)\right)=c_{k}$ for some $k$. Additionally,
$\left|T^{-1}\left(c_{k}\right)\right|,k=1,...,K$ are proportional
to $\pi_{k}^m,k=1,...,K$. Denote $\sigma:\left\{ 1,...,N\right\} \goto\left\{ 1,...,K\right\} $
such that $T\left(f_{e}^m\left(x_{n}\right)\right)=c_{\sigma\left(n\right)},\forall i=1,...,N$,
we have $\sigma\in\Sigma_{\pi}$. It also follows that 
\[
\mathcal{W}_{\rho_{z}}\left(\frac{1}{N}\sum_{n=1}^{N}\delta_{f_{e}^m\left(x_{n}\right)},\sum_{k=1}^{K}\pi_{k}^m\delta_{c_{k}}\right)=\frac{1}{N}\min_{\sigma\in\Sigma_{\pi}}\sum_{n=1}^{N}\rho_{z}\left(f_{e}^m\left(x_{n}\right),c_{\sigma\left(n\right)}\right).
\]

\section{Practical Algorithm for VQ-WAE}
\label{alg_learn_pi_entropic_dual}

We first re-introduce the entropic regularized dual form of optimal transport by \citep{genevay2016stochastic} which 
enables the application of optimal transport in machine
learning and deep learning:

\begin{equation}
    \mathcal{W}_{d}^{\epsilon}\left ( \mathbb{Q},\mathbb{P} \right ):=\min_{\gamma \in \Gamma( \mathbb{Q},\mathbb{P} )} \left\{ \mathbb{E}_{(x,y)\sim\gamma} \left [ d(x,y) \right ] + \epsilon D_{KL}(\gamma\parallel \mathbb{Q}\otimes \mathbb{P}) \right\} 
    \label{apd:entropic_reg}
\end{equation}

where $\epsilon$ is the regularization rate,$D_{KL}(\cdot \parallel \cdot)$ is the Kullback-Leibler (KL) divergence, an $\mathbb{Q}\otimes \mathbb{P}$ represents the specific coupling in which $\mathbb{Q}$ and $ \mathbb{P}$ are independent. 

Second, using the Fenchel-Rockafellar theorem, they obtained the following dual form w.r.t. the potential $\phi$:

\begin{equation}
\mathcal{W}_{d}^{\epsilon}\left(\mathbb{Q},\mathbb{P} \right )=\max_{\phi}\left \{ \mathbb{E}_\mathbb{Q}[\phi^c_\epsilon(x)]+\mathbb{E}_\mathbb{P}[\phi(y)] \right \}
\label{apd:entropic_dual_form}
\end{equation}

where $\phi^c_\epsilon(x)=-\epsilon\log\left ( \mathbb{E}_{\mathbb{P}}\left [ \exp\left \{ \frac{-d(x,y)+\phi(y))}{\epsilon} \right \} \right ] \right )$.

\label{apd:practical_algorithm}

We now present how to develop a practical method for our VQ-WAE by entropic regularized dual form (\ref{apd:entropic_dual_form}). We rewrite our objective function:

\begin{equation}
\min_{C,\pi, f_{d},f_{e}}
\begin{Bmatrix}
\mathbb{E}_{x\sim\mathbb{P}_{x}}\left[d_{x}\left(f_{d}\left(Q_C\left(f_{e}\left(x\right)\right)\right),x\right)\right]   \\
+ \frac{\lambda}{M} \times\sum_{m=1}^{M} \mathcal{W}_{\rho_{z}}\left(f_{e}^{m}\#\mathbb{P}_{x},\mathbb{P}_{c,\pi^{m}}\right) \\
+\lambda_r \sum_{m=1}^{M}D_{KL}(\pi^m, \mathcal{U}_K)
\end{Bmatrix}
\end{equation}
where $\lambda, \lambda_r > 0$ are two trade-off parameters and  and $\mathcal{U}_K=\left[\frac{1}{K}\right]_{K}$.

To learn the weights $\pi$,
we parameterize $\pi^m = \pi^m(\beta^m) = \text{softmax}(\beta^m), m=1,\dots,M$ with $\beta^m \in \mathbb{R}^K$. At each iteration, we sample a mini-batch $x_1,...,x_B$
and then solve the above OP by updating $f_{d},f_{e}$ and $C,\beta^{1..M}$
based on this mini-batch as follows. Let us denote 
$$\mathbb{P}_{B}=\frac{1}{B}\sum_{i=1}^{B}\delta_{x_{i}}$$
as the empirical distribution over the current batch.

For each mini-batch, we replace $\mathcal{W}_{\rho_{z}}\left(f_{e}^{m}\#\mathbb{P}_{x},\mathbb{P}_{c,\pi^{m}}\right)$ by $\mathcal{W}_{\rho_{z}}\left(f_{e}^{m}\#\mathbb{P}_{B},\mathbb{P}_{c,\pi^{m}}\right)$
and approximate it with
entropic regularized duality form $\mathcal{R}^m_{WS}$ (see Eq. (\ref{apd:entropic_dual_form})) as follows:

\begin{equation} \mathcal{R}^m_{WS}
=
\max_{\phi^m}\left \{ \frac{1}{B}\sum_{i=1}^{B} \left [ -\epsilon\log\left ( \sum_{k=1}^{K}\pi^m_k\left [ \exp\left \{ \frac{-\rho_z(f_e^m(x_i),c_k)+\phi^m\left ( c_k\right )}{\epsilon} \right \} \right ] \right ) \right ]+\sum_{k=1}^{K}\pi^m_k\phi^m(c_k) \right \}
\end{equation}

where $\phi^m$ is a neural net named Kantorovich potential network.

Finally,
we update $f_{d},f_{e},C,\beta^{1:M}$ by solving for each mini-batch:
\begin{equation}
\min_{C,\beta^{1...M}}\min_{f_{d},f_{e}}\max_{\phi^{1...M}}\left\{ \frac{1}{B}\sum_{i=1}^{B}d_{x}\left(f_{d}\left(Q\left(f_{e}\left(x_{i}\right)\right)\right)\right)
+
\sum_{m=1}^{M}
\left(
\frac{\lambda}{M}\mathcal{R}^m_{WS}
+ \lambda_r D_{KL}\left(\pi^{m}\left(\beta^{m}\right), \mathcal{U}_K\right)
\right)\right\} .\label{eq:appendix_final_opt}
\end{equation}

Note that we can optimize $M$ WS distances $\mathcal{W}^{\epsilon}_{d_{z}}\left(f_{e}^{m}\#\mathbb{P}_{N},\mathbb{P}_{c,\pi^{m}}\right)$ in parallel by matrix computation from current deep learning framework.

\section{Analysis of $\lambda$ and $\lambda_r$}
\label{sec:KL_analysis}

\begin{table}[h!]
    \centering
    \caption{Reconstruction performance of VQ-WAE with different $\lambda$ values on CIFAR10 dataset.}
    \begin{tabular}{llllrrr}
        \toprule 
Model & $\lambda_r$ &  $\lambda$  & rFID $\downarrow$ & Perplexity $\uparrow$
\tabularnewline 
\midrule 
\multirow{6}{*}{VQ-WAE} & & $1e^{-2}$ & 55.82 & 504.5 \tabularnewline
 &1.0  & $1e^{-3}$ & \textbf{54.30} & 497.3 \tabularnewline
&& $1e^{-4}$ &  58.96 & \textbf{507.9}
\tabularnewline
 \cmidrule{2-5}
& & $1e^{-2}$ & 68.99 & 445.8 \tabularnewline
 & 0.0  & $1e^{-3}$ & 57.49 & 456.5 \tabularnewline
& & $1e^{-4}$ & 58.17 & 467.8
\tabularnewline
\midrule
VQ-VAE  &&& 77.3 & 69.8  \tabularnewline
 SQ-VAE &&  & 55.4 & 434.8 \tabularnewline
        \bottomrule
    \end{tabular}
    \label{tab:lambda}
\end{table}

In this section, we provide further elaboration on the rationale behind employing regularization on $\pi^m$ to enforce a uniform distribution, as denoted by the third term in objective~\ref{eq:KL_regularization_OP}. The first motivation stems from the desire to ensure the utilization of every discrete codeword. Specifically, we have observed that in the absence of KL regularization (i.e., $\lambda_r=0.0$), the complexity can be reduced. This reduction occurs because during the optimization of $\left \{ \pi^m \right \}_{m=1}^M$, certain $\pi^{m}_k$ values can significantly decrease and converge to zero, resulting in low usage of certain codewords.

\begin{figure}[h!]

\centering
\includegraphics[width=1.0\textwidth]{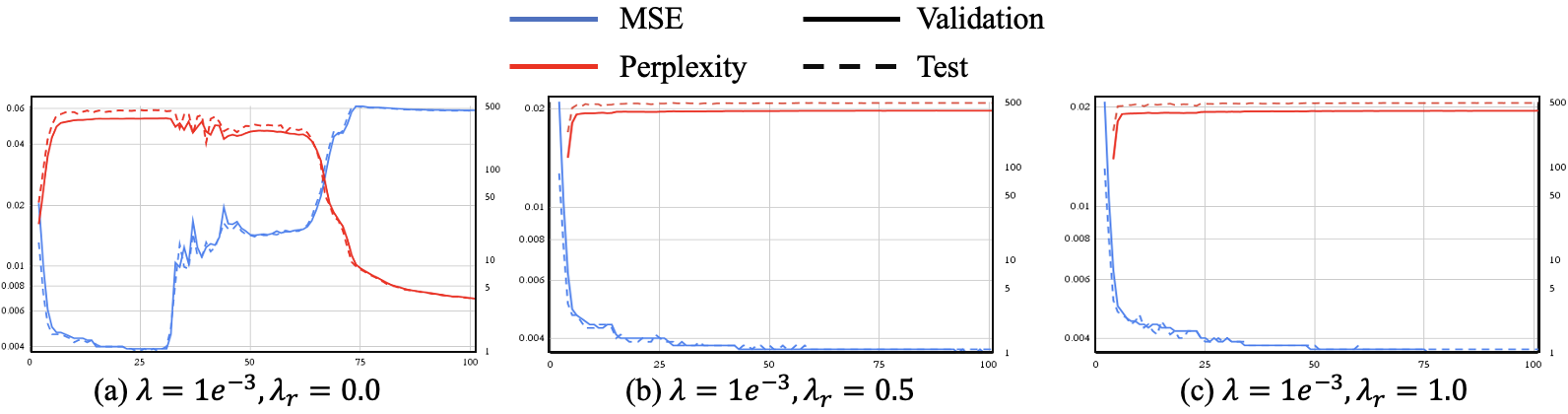}
\caption{Training and Validation curve of CIFAR10 with different $\lambda_r$.}
\label{fig:training_curve}
\end{figure}

Secondly, we have observed that training VQ-WAE without KL-regularization leads to divergence after convergence (Figure~\ref{fig:training_curve}.a). However, the addition of a small KL-regularization term not only enhances model performance but also stabilizes the training process (Figure~\ref{fig:training_curve}.b and Figure~\ref{fig:training_curve}.c). Furthermore, the results presented in Table~\ref{tab:lambda} demonstrate that in the absence of KL-regularization ($\lambda_r=0.0$), performance exhibits significant variability when the value of $\lambda$ changes. This finding suggests that incorporating the KL-regularization term reduces the model's sensitivity to variations in $\lambda$. Additionally, we report the performance of VQ-WAE on CIFAR10 with a fixed $\pi$ assumed to be a uniform distribution (Table~\ref{tab:lambda}). The findings indicate that extremely high perplexity can have a detrimental impact on performance.


\section{Experimental Settings}
\label{apd:expsetting}

\subsection{VQ-model}
 \textbf{Implementation}: For fair comparison, we utilize the same framework architecture and hyper-parameters for both VQ-VAE and VQ-WAE. 
Specifically, we construct the VQ-VAE and VQ-WAE models as follows:

\begin{itemize}
    \item For CIFAR10, MNIST and SVHN datasets, the models have an encoder with two convolutional layers of stride 2 and filter size of 4 × 4 with ReLU activation, followed by 2 residual blocks, which contained a 3 × 3, stride 1 convolutional layer with ReLU activation followed by a 1 × 1 convolution. The decoder was similar, with two of these residual blocks followed by two deconvolutional layers.
    
    \item For CelebA dataset, the models have an encoder with two convolutional layers of stride 2 and filter size of 4 × 4 with ReLU activation, followed by 6 residual blocks, which contained a 3 × 3, stride 1 convolutional layer with ReLU activation followed by a 1 × 1 convolution. The decoder was similar, with two of these residual blocks followed by two deconvolutional layers.
    
    \item For high-quality image dataset FFHQ, we utilize the well-known VQGAN framework \cite{esser2021taming} as the baseline.
\end{itemize}

\textbf{Hyper-parameters}: Following \citep{takida22a}, we adopt the \textit{adam optimizer} for training with: \textit{learning-rate} is $e^{-3}$,  \textit{batch-size} of $32$,  \textit{embedding dimension} of $64$ and \textit{codebook size} $\left |C \right |=512$ for all datasets except FFHQ with \textit{embedding dimension} of $256$ and $\left | C \right|=1024$. Finally, we train model for CIFAR10, MNIST, SVHN, FFHQ in 100 epoches and for CelebA in 70 epoches respectively.

\textbf{Time Complexity:} We report extra computation required by VQ-WAE on CIFAR dataset.
Note that we need to trains a kantorovich network to estimate
the empirical Wasserstein distance which take extra computation for training. In our experiments, the kantorovich network is designed with a hidden layer of $M \times 64$ nodes where $M$ is the number of components of a latent while $64$ is the embedding dimension. The training steps
\textit{$\phi$-iteration is set to 5} which is chosen for fast computation and sufficient optimization.
Precisely on the system of a GPU NVIDIA Tesla V100 with dual CPUs Intel Xeon E5-2698 v4, training VQ-WAE
takes about \textit{$64$ seconds} for one epoch on CIFAR10 dataset, while training a standard VQ-VAE only takes approximately \textit{$40$ seconds} for one epoch. For inference, both methods take the same time.

\subsection{Generation model}
\textbf{Implementation}: It is worth to noting that we employ the codebooks learned from reported VQ-models to extract codeword indices and we use the same model for generation for both VQ-VAE and WQ-VAE. 
\begin{itemize}
    \item CIFAR10, MNIST and SVHN contain the images of shape $(32,32,3)$ and latent of shape $(8,8,1)$, we feed PixelCNN over the "pixel" values of the $8\times 8$ $1$-channel latent space.
    
    \item CelebA contains the images of shape $(64,64,3)$ and latent of shape $(16,16,1)$, we feed PixelCNN over the "pixel" values of the $16\times 16$ $1$-channel latent space.
\end{itemize}

\textbf{Hyper-parameters}: we adopt the \textit{adam optimizer} for training with: \textit{learning-rate} is $3e^{-4}$,  \textit{batch-size} of $32$.



\end{document}